%% file: cikm-main.tex
\documentclass[sigconf]{acmart}

\usepackage{booktabs} % For formal tables

\copyrightyear{2018} 
\acmYear{2018} 
\setcopyright{acmcopyright}
\acmConference[CIKM '18]{2018 ACM Conference on Information and Knowledge Management}{October 22--26, 2018}{Torino, Italy}
\acmBooktitle{2018 ACM Conference on Information and Knowledge Management (CIKM'18), October 22--26, 2018, Torino, Italy}
\acmPrice{15.00} 
\acmDOI{10.1145/3269206.3271798}
\acmISBN{978-1-4503-6014-2/18/10}

\usepackage{graphicx}
\usepackage{balance} 

\usepackage{alltt} 

\input{header}

\begin{document}
\title{Detecting Outliers in Data with Correlated Measures}

%\begin{comment}
\author{Yu-Hsuan Kuo}
%\authornote{Dr.~Trovato insisted his name be first.}
%\orcid{1234-5678-9012}
\affiliation{
\institution{Dept. of Computer Science \& Engineering\\
Pennsylvania State University}
%  \streetaddress{P.O. Box 1212}
%  \city{Dublin}
%  \state{Ohio}
% \postcode{43017-6221}
}
\email{yzk5145@cse.psu.edu}

\author{Zhenhui Li}
%\authornote{The secretary disavows any knowledge of this author's actions.}
\affiliation{%
  \institution{College of Information Sciences \& Technology\\
  Pennsylvania State University}
 % \streetaddress{P.O. Box 1212}
 % \city{Dublin}
 % \state{Ohio}
 % \postcode{43017-6221}
}
\email{ jessieli@ist.psu.edu}

\author{Daniel Kifer}
\affiliation{
\institution{Dept. of Computer Science \& Engineering\\
Pennsylvania State University}
}
\email{dkifer@cse.psu.edu}

\input{0abstract2_detector}


\maketitle

\input{1intro2_detector}

\input{5related_works_detector}

\input{2problem_def_detector}

\input{3method-noise_v1}

\input{6experiment_detector}

\input{7conclusion_detector}

\section{Acknowledgments}
The work was partially supported by NSF grants 1054389, 1544455, 1702760, 1652525. The views in this paper are those of the authors, and do not necessarily represent the funding institutes.

\balance
\bibliographystyle{ACM-Reference-Format}
\bibliography{ref_outlier}%\bibliography{sample-bibliography}

\end{document}

%% file: header.tex
\usepackage{amssymb,amsmath,epsfig,graphics,enumerate,float,subfigure,verbatim,xspace,color, wrapfig,url}
\usepackage{hyperref} %hyperlink
\usepackage{multirow}
\usepackage{pbox}

\usepackage{algorithm}
\usepackage{algpseudocode} %add this then i can show the \EndFor

\usepackage{verbatim} %for begin comment
\usepackage{booktabs}  %for top rule bottom rule
\usepackage{commath} %math absolute value
\usepackage{gensymb} %for degree
\usepackage{tablefootnote}
\usepackage{tabularx}
\usepackage{listings}% http://ctan.org/pkg/listings
\usepackage{alltt}

\usepackage[capitalise]{cleveref}

\usepackage{graphicx}
\usepackage{afterpage}

\usepackage{enumitem}%reduce some of the indent in enumerate and itemize environments or spacing between the items in enumerate and itemize.

\algdef{SE}[DOWHILE]{Do}{doWhile}{\algorithmicdo}[1]{\algorithmicwhile\ #1} %do while

\newcolumntype{L}[1]{>{\raggedright\arraybackslash}p{#1}} %for content in the table
\newcolumntype{C}[1]{>{\centering\arraybackslash}p{#1}}
\newcolumntype{R}[1]{>{\raggedleft\arraybackslash}p{#1}}

\newcommand{\SysName}{\texttt{Doc}\xspace}  %name:  odie (outlier detection interpretable explanation), cod (correlation outlier detection), 

\DeclareMathOperator{\database}{\mathcal{I}}

\newcommand{\nop}[1]{}

%% file: 0abstract2_detector.tex
\begin{abstract}
Advances in sensor technology have enabled the collection of large-scale datasets. Such datasets can be extremely noisy and often contain a significant amount of outliers that result from sensor malfunction or human operation faults. In order to utilize such data for real-world applications, it is critical to detect outliers so that models built from these datasets will not be skewed by outliers.

 In this paper, we propose a new outlier detection method that utilizes the correlations in the data (e.g., taxi trip distance vs. trip time). Different from existing outlier detection methods, we build a robust regression model that explicitly models the outliers and detects outliers simultaneously with the model fitting. %In addition, we propose a systematic approach to explain outliers. While previous studies typically learn subspaces that separate outliers from non-outliers, our method can learn more sophisticated but meaningful rules by considering complex operations of attributes based on the attribute units.

We validate our approach on real-world datasets against methods specifically designed for each dataset as well as the state of the art outlier detectors. Our outlier detection method achieves better performances, demonstrating the robustness and generality of our method. Last, we report interesting case studies on some outliers that result from atypical events.%More importantly, our outlier interpreter generates more meaningful rules to summarize and explain the outliers and show insights into outliers. 
\end{abstract}

%% file: 1intro2_detector.tex
% !TEX root = cikm-main.tex
\section{Introduction}

With the development in sensor technology, increasing amount of data collected from sensors become publicly available.  Analyzing such data could benefit many applications such as smart city, transportation, and sustainability. For example, New York City (NYC) has released a massive taxi data set~\cite{nyctaxi} including information such as pickup and dropoff locations and time, trip cost, and trip distance. Such data have been used for studies such as characterizing urban dynamics~\cite{qian2015characterizing}, detecting events in city~\cite{zheng2015detecting}, and estimating travel time~\cite{wang2016simple}.

In these large-scale sensor datasets, there could be a significant amount of outliers due to sensor malfunction or human operation faults. For example, in NYC taxi data, we have observed trips with extremely long moving distances but unreasonably low trip fares. There are also trips with short displacements between pickup and dropoff locations but have a long trip distance. In a recent work on travel time estimation~\cite{wang2016simple}, Wang et al. found that such outliers in the original datasets can break effective travel time estimation methods.

\begin{figure}[htb]%h: here, t: top, b: bottom
\centering
\includegraphics[width=0.43\textwidth]{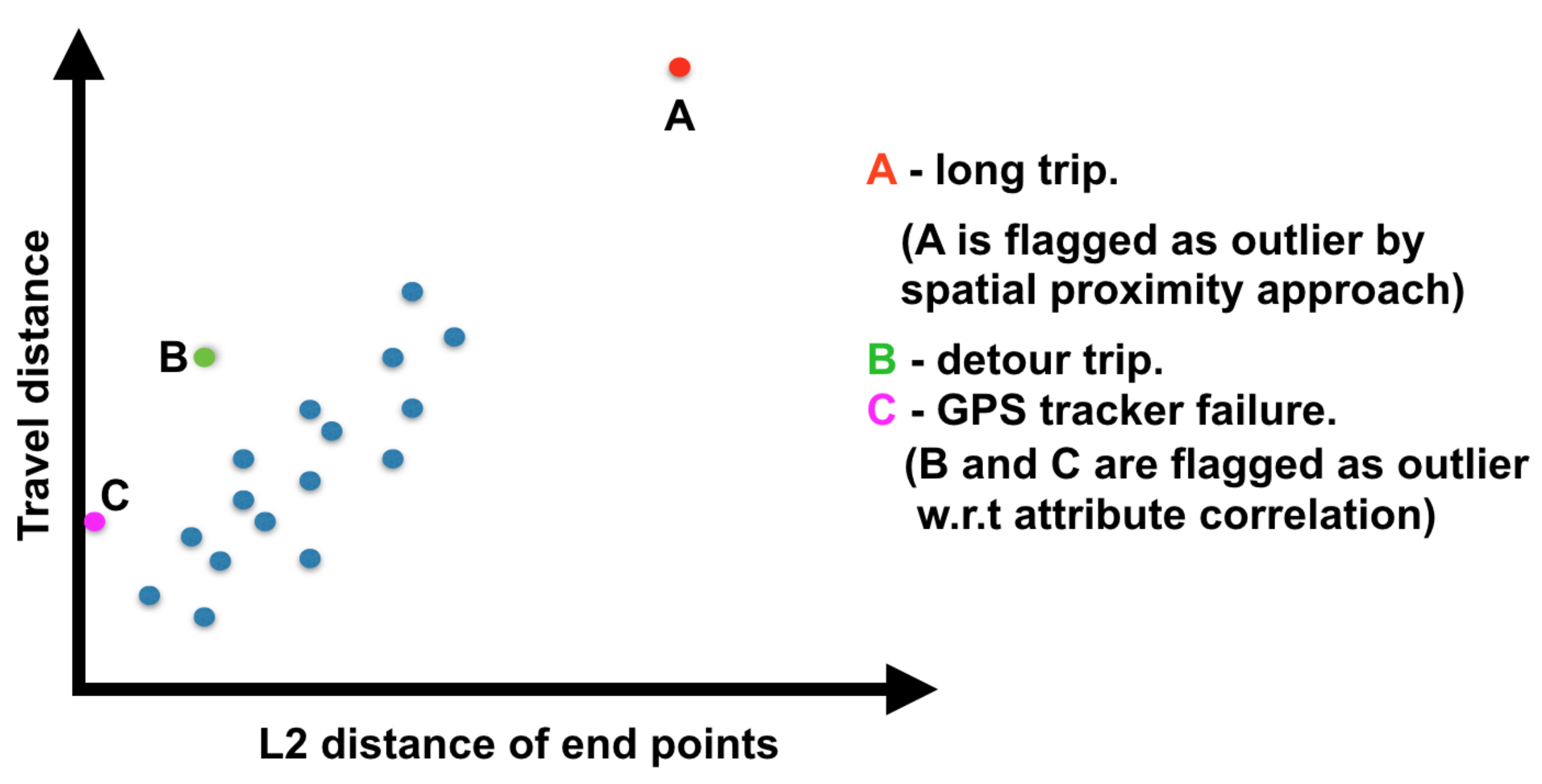}%{figs/eg_legend.png}
%\vspace{-5pt}
\caption{Taxi trip example: suspicious outlying trip}
\label{fig:eg}
\end{figure}

There have been many methods proposed in literature on outlier detection~\cite{chandola2009anomaly}. Typical outlier detection methods define a sample as an outlier if it significantly deviates from other data samples. However, such definition may not apply in our case. Consider an example shown in Figure~\ref{fig:eg}. There could  be many interpretations of what is an outlier in this figure. One possibility is point A is an outlier while points B and C are more likely to be labeled as normal points based on the spatial proximity of every datum to its neighbors. However, another possibility is sample A could be a long but normal trip because the ratio between travel distance and L2 distance between end points is within the normal range. On the other hand, sample B and sample C, even though being closer to other data samples, could be outliers. Sample B could be a trip with detour because the travel distance is much longer than L2 distance between end points. Sample C has a nearly zero L2 distance (i.e., the same pickup and dropoff locations), which could be an outlier due to sensor malfunction.

\begin{figure}[htb]%h: here, t: top, b: bottom
\centering
\includegraphics[width=0.43\textwidth]{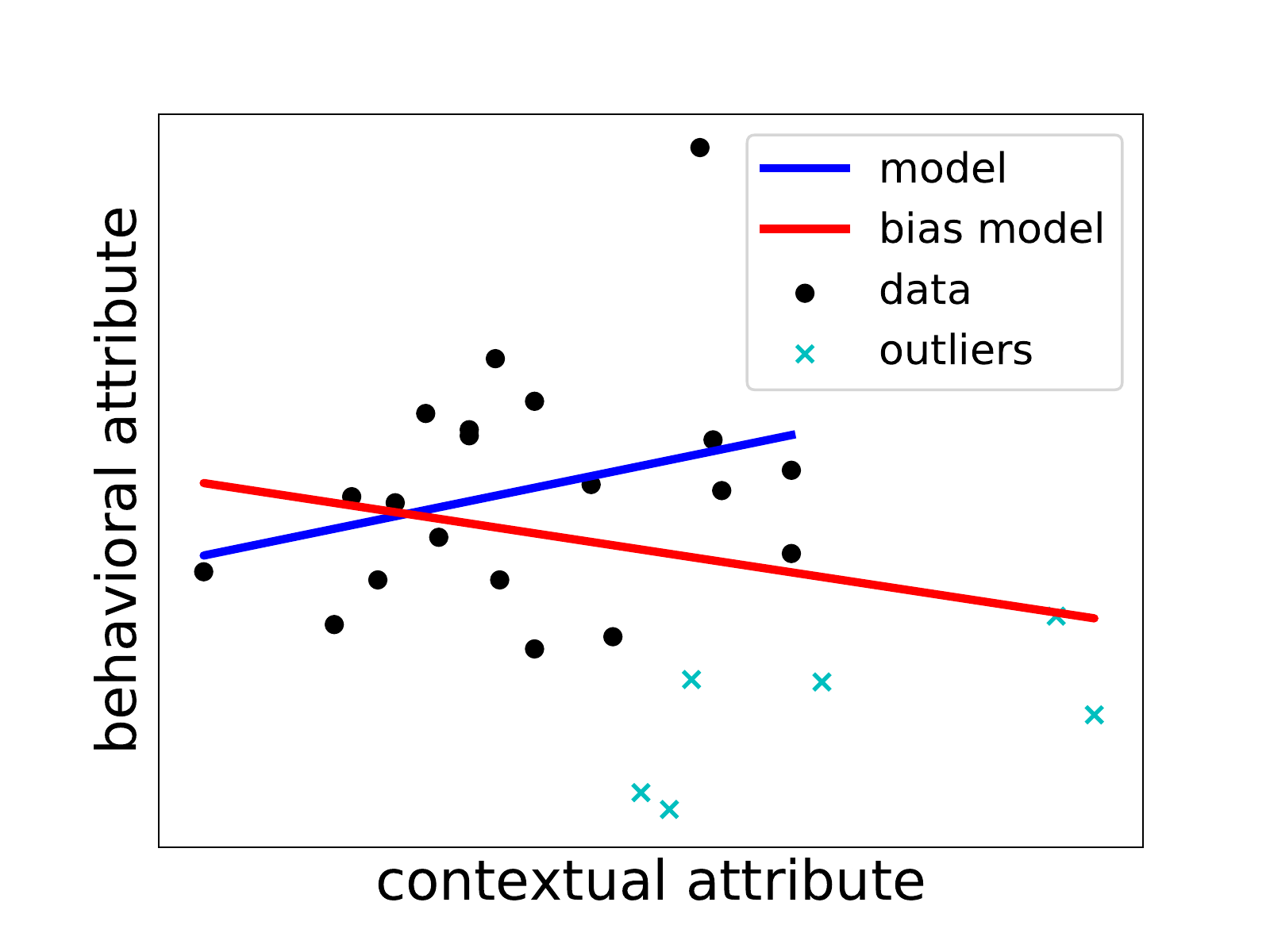}%
\vspace{-8pt}
\caption{Biased Model Illustration}
\label{fig:bias}
\end{figure}

% To avoid dealing with subjectivity, Song et. al~\cite{song2007conditional} proposed \emph{contextual outlier detection}. This model based approach learns a model over the data, and data points that are poorly modeled are labeled as outliers. 
Motivated by the observations on real-world data, we detect outliers based on empirical correlations of attributes, which is close to the \emph{contextual outlier detection} proposed by Song et. al.~\cite{song2007conditional}. For example, we expect correlations between attributes trip time and trip distance in taxi data, and between voltage and temperature in CPU sensor data. If the attributes of a data sample significantly deviate from expected correlations, this data sample is likely to be an anomaly. Domain experts can specify correlation templates so that the definition of an outlier can be customized to the application. We propose a robust regression model that explicitly models the non-outliers and outliers. We feed the algorithm domain knowledge about correlations (e.g., the fact that trip time should be predictable from trip distance \& time of day) and it learns how to model them (e.g., how to predict trip time from trip distance and time of day). The model is robust (so outliers do not skew the model parameters) and automatically generates a probability for each data sample being as an outlier and also automatically generates a cut-off threshold on probabilities for outliers. 

%It is worth noting that, 
In literature, there is a series of contextual outlier detection methods that use the correlation between contextual attributes and behavioral attributes to detect outliers~\cite{song2007conditional, hong2015mcode, liang2016robust}.
One problem with contextual outlier detection is that outliers can bias a model that is learned from noisy data. To the best of our knowledge, prior work on contextual outlier detection did not consider this issue. The biased model could end up marking outliers as non-outliers and non-outliers as outliers. Take \Cref{fig:bias} as an example. The blue line indicates a model that would have been learned if it was trained on clean data. However, because clean data is not available, contextual outlier detection trains on noisy data. The red line shows the result. To address this problem, we propose a regression model that explicitly models for outliers and non-outliers.

We conduct experiments on four real-world datasets and demonstrate the effectiveness of our proposed method with comparison to classical regression methods and five existing outlier detection algorithms.  With the help of our model, the root cause of outliers can be identified. For example, in the taxi dataset, we found that many outliers are from sensors produced by a certain manufacturer. We report case studies to support our detected outliers and provide insights into the data that can then be used to study new phenomena or devise ways to improve sensors reliability.

In summary, our key contributions are:

\begin{itemize}[leftmargin=*,topsep=2pt,itemsep=1pt] 
\item We propose an outlier detection method that utilizes correlations between attributes. Such correlations can be specified by domain experts depending on the application. Different from existing work, our method is a robust regression model that explicitly considers outliers and automatically learns the probability for a data sample being an outlier. It intrinsically generates the thresholds for classification while being robust to parameter skewed by outliers, which is a common problem with other approaches. 

\item We conduct rigorous experiments on real-world datasets. For these datasets with missing ground truth, human annotation system is used to obtain labels. We design the machine learning task to show that outliers may bias the model trained on unsanitized dataset. We also inject synthetic outliers to validate the model's robustness to different types of outliers.
\item We compare our approach against five recent outlier detectors (including other contextual outlier detection algorithms). Our method significantly outperformed competing methods and continues to perform well even in extremely noisy datasets (which are common in big data obtained from sensor measurements).

\end{itemize}

The rest of the paper is organized as follows. Related work is discussed in \cref{sec:related}.  \cref{sec:problem} describes the system overview. We present our outlier model in \cref{sec:noise}. % and rule-based outlier explainer in  \cref{sec:rules}. 
We then empirically evaluate our methods in \cref{sec:exp}. We present conclusions in  \cref{sec:conclusion}.

%% file: 5related_works_detector.tex
% !TEX root = cikm-main.tex

\section{Related Work}
\label{sec:related}

%\noindent \textbf{Outlier Detection.}

We outline the progress related to two categories:  \emph{unsupervised outlier detection for numerical datasets} and \emph{contextual outlier detection}. %and \emph{outlier detection for vehicle trip data}.
 
 \subsection{Unsupervised Outlier Detection}
Typical unsupervised outlier detection methods aim to find data samples that are significantly different from other samples. %The difference is measured either based on the distance~\cite{Knorr:1998:AMD:645924.671334,ramaswamy2000efficient,sugiyama2013rapid} or the density~\cite{papadimitriou2003loci,breunig2000lof,Jin:2001:MTL:502512.502554}. Some studies assume that data is generated from an underlying statistical distribution~\cite{yamanishi2000line} and outliers can be defined by the distribution.  However, all these studies do not utilize the correlations of attributes for outlier detection.
 Yamanishi et al. \cite{yamanishi2000line} assume that data is generated from an underlying statistical distribution. The notion of outlier is captured by a strong deviation from the presumed data dependent probabilistic distribution. 
 In distance-based outlier work \cite{Knorr:1998:AMD:645924.671334,ramaswamy2000efficient,sugiyama2013rapid}, they measure the distance of a data point to its neighbors. The assumption is that normal objects have a dense neighborhood, thus the outlier is the one furthest from its neighbors. 
 Similar approaches using the spatial proximity are density-based \cite{papadimitriou2003loci,breunig2000lof,Jin:2001:MTL:502512.502554}. These works adopt the concept of neighbors by measuring the density around a given datum as well as its neighborhood. Breunig et al. \cite{breunig2000lof} introduce a local outlier factor (LOF) for each object in the dataset, indicating its degree of outlierness. The outlier factor is \emph{local} in the sense that the degree depends on how isolated the object is with respect to only neighboring points. 
 These outlier algorithms consider different characteristics and properties of anomalous objects in a dataset. These outlying properties can vary largely on the type of data and the application domain for which the algorithm is being developed. However, all these studies do not consider the outlying behavior with respect to a given context, assuming every attribute contributes equally to the feature vector.

\subsection{Contextual Outlier Detection}
Another line of works related to our correlation templates is contextual/conditional outlier detection where one set of attributes defines the context and the other set is examined for unusual behaviors. Song et al. \cite{song2007conditional} propose conditional anomaly detection that takes into account the user-specified environmental variables. Hong et al. \cite{hong2015mcode} model the data distribution by multivariate function and transform the output space into a new unconditional space. %develop conditional outlier detection method in which data objects are associated with multi-dimensional binary outputs (responses)  
Lang et al. \cite{liang2016robust} model the relationship of behavioral attributes and contextual attributes from local perspectives  (i.e., contextual neighbors)  as well as global perspectives. However, none of these works build their models under the awareness/assumption of outlier and thus the training process is limited to clean data.  

There are also contextual outlier detection for graphs \cite{valko2011conditional, wang2009discovering}  and categorical data \cite{tang2015mining}. Valko et al. \cite{valko2011conditional} proposed a non-parametric graph-based algorithm to detect conditional anomalies. However it assumes the labeled training set is available. Wang et al. \cite{wang2009discovering} address the problem of detecting contextual outliers in graphs using random walk. Tang et al.  \cite{tang2015mining} identify contextual outliers on categorical relational
data by leveraging data cube computation techniques. But they are not applicable to numerical data used in our work.

%% file: 2problem_def_detector.tex
% !TEX root =cikm-main.tex
\section{Notations and System Overview}%\section{Problem Definition}
\label{sec:problem}

%\Note{Instead, it should introduce notation for a dataset, attributes, how to get record i and attribute j of record i. why?}
%A dataset $\database$ is a collection of $n$ records with $m$ attributes $\mathcal{A} = \{x_1, x_2, \dots x_m \}$. Let $x_{mn}$ denote the attribute $m$ of record $n$. Let $\vec{z}$ be a feature vector where each element could from $\mathcal{A}$ or other external variables. Let $g(x_i, \vec{z})$ be a function which transforms $x_i$ and $\vec{z}$ into an approximate linear relationship. Given the domain knowledge that $\database$ preserves $C$ correlations $Corr =$ $\{ g(x_1, \vec{z_1}), \dots,$ $g(x_C, \vec{z_C})\}$, the inputs to the outlier detector are $\database$ and user-specified $Corr$. %Note that $A_i$ (or $A_j$) can be the attribute or derived from the attributes (e.g., the L2 norm of attributes $x_1, x_2$:  $A_i = || x_1 - x_2 ||_2$.)  

A dataset $\database$ is a collection of $n$ records $\{\vec{z}_1, \dots, \vec{z}_n\}$  where each $\vec{z}_i$ has $m$ attributes $\vec{z}_i[1], \dots, \vec{z}_i[m]$. %Each attribute $j$ can be associated with a unit of measurement $U_j$ (e.g., feet, seconds, etc.). $U_j$ can be null (e.g., attribute $j$ is a manufacturer).
 
A \emph{correlation template} is a pair $(j, S)$ where $j$ is a behavior attribute and $S$ $\subseteq\set{1,\dots, m}$ is a set of contextual attributes. This means that the value $\vec{z}_i[j]$ can be predicted from attributes $\vec{z}_i[s]$ for $s \in S$. 

To avoid heavy use of sub-subscripts, we will also use the following renaming. For a correlation $(j, S)$, we set $y_i$ to be $\vec{z}_i[j]$ and $\vec{x}_i$ to be the vector of the attribute values in $S$ (i.e. $\vec{x}_i = [\vec{z}_i[s] \text{ for } s \in S]$).

\begin{figure}[t!]
\centering
\includegraphics[width=0.43\textwidth]{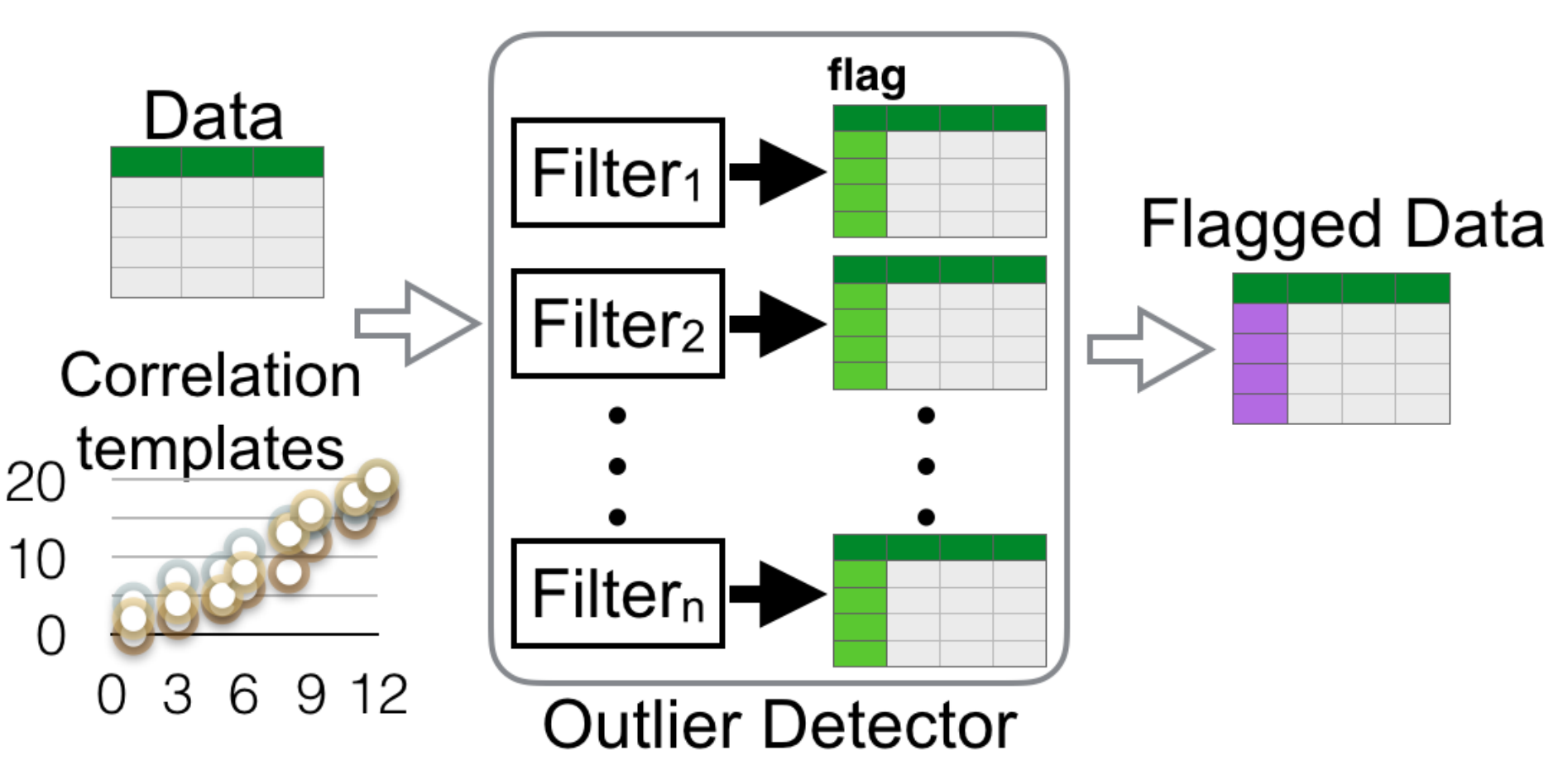}
\vspace{-4pt}
\caption{\SysName System Overview}
\label{fig:SysName-detector}
\end{figure}

An overview of the outlier detector, called $\SysName$, is shown in \cref{fig:SysName-detector}. It contains an outlier detector that flags suspicious records.% and an outlier explainer that generates rules.

The inputs to the outlier detector are $\database$ and a set $Corr$ of $C$ correlation templates $Corr= \{ (j_c, S_c)\}_{c=1}^{C}$. In different applications, some attributes $j$ are usually associated with outlier behavior; but if its relevant attributes $S$ are not specified by domain experts, the system will take the rest of attributes as $S$, serving as the context of the behavior.

%We define the \textbf{filter} as a model to learn a correlation $g(x_i, \vec{z})$.  Each filter $i$ provides each record an outlier probability. It sorts the dataset by such outlier probability and the estimated fraction of top rank records are flagged as outlier where the flag of an outlier from filter $i$ is $\ell_i = 1$.  The outlier detector contains a sequence of filters and outputs the final flagged dataset $\tilde{\database}$  where it takes a record into the outlier set if its $\ell_1 \vee \ell_2 \vee \cdots \vee \ell_i = 1$.  

In the outlier detector, a \emph{filter} is a model that learns how to predict $\vec{z}[j]$ from the $\vec{z}[s]$ for $s\in S$. The goal of each filter is to assign a score $t_i$  to every record indicating its estimated probability that the record is an outlier (this is described in Section \ref{sec:noise}). Higher score implies its higher probability of being an outlier. The expected number of outliers $K$ is the sum of these scores $t_i$, and the top $K$ records are flagged as outliers by the filter. When using multiple filters, a record is marked as an outlier if at least one filter marks it as an outlier. We average outlier scores returned from multiple filters as an overall outlier score of a record. The result is a dataset $\tilde{\database}$ in which every record $\vec{z}_i$ has a flag $\ell_i$ indicating whether it should be considered an outlier ($\ell_i=1$) or not ($\ell_i=0$).

The summary of notations is in \cref{tab:notation}.

\vspace{-10pt}
\begin{table}[htb] %h here , t:top , b, bottom
\center
\caption{Notations}
\begin{tabular}{ | c | c | }
%\toprule
\hline 
  $(j, S)$     & Correlation template taken by a filter     \\
\hline
$y_i$ =  $\vec{z}_i[j]$ & Behavioral attribute $j$ value of record $\vec{z_i}$\\
\hline
 $\vec{x}_i = $ &  \multirow{2}{*}{Contextual attributes $S$ of record $\vec{z_i}$} \\
$ [\vec{z}_i[s] \text{ for } s \in S]$				& \\
\hline
\multirow{2}{*}{$t_i$} & Outlier score of record $z_i$ provided \\
& by a filter.\\
\hline
$K$ &\# records flagged as outliers by a filter.\\
\hline
$\ell_i$ & Outlier flag of record $\vec{z_i}$\\
\hline
%\bottomrule
\end{tabular}
\label{tab:notation}
\end{table}

%%%%%%%%%%explainer%%%%%%%%%%
\nop{
Given $\tilde{\database}$ and a type-system over units (described in Section \ref{sec:rules}), the system then searches for rules that characterize the flagged outliers (in Section \ref{sec:rules}).

\medskip
%Our goal is to identify outliers including the erroneous data and the real outliers. Then the system returns rules that give information about the identified outliers. Our system $\SysName$ overview is in \cref{fig:SysName} where the filter is introduced in \cref{sec:noise} and the explanation framework is in \cref{sec:rules}.

%Our system $\SysName$ overview is in \cref{fig:SysName} where it is comprised of outlier detector and outlier explainer. The outlier detector takes the data and observed correlations (i.e., L2 distance is positively correlated with actual distance) as input and produces flagged outliers as output. The explainer takes flagged result and data features (such as trip fare, pickup neighborhood, etc.) to extract outlier rules.

\begin{figure}[t!]
\centering
\includegraphics[width=0.40\textwidth]{figs/SysName.pdf}
\vspace{-4pt}
\caption{\SysName System Overview}
\label{fig:SysName}
\end{figure}
}

%% file: 3method-noise_v1.tex
%!TEX root = oddd.tex
% !TEX root = outlier_main.tex

\section{Outlier Detector}
\label{sec:noise}

The job of the outlier detector is to take each correlation template $(j, S)$ and learn a model that, for each record $\vec{z}_i$, can  predict $\vec{z}_i[j]$ from the attributes $\vec{z}_i[s]$ for $s\in S$. It then assigns an outlier score $t_i$ to each record $\vec{z}_i$. This score is the estimated probability that the record is an outlier and is based on how much the actual value $\vec{z}_i$ deviates from its prediction.

We do this by modeling the prediction error as a mixture of light-tailed distributions (for non-outliers) and heavy-tailed distributions (for outliers). Similar noise mixtures are used in robust statistics
\cite{Carrillo:2010:GCD:1928599.1957893,rosa2003robust,west1987scale,huber1964robust}, and typically M-estimators or MCMC inference are used to find model parameters. Instead, we specifically use a variant of expectation-maximization (EM) \cite{DEMP1977} because it produces variables that, as explained in Section \ref{subsec:outlier}, can be interpreted as outlier probabilities $t_i$. Indeed, we are more interested in these $t_i$ than in the model parameters themselves.

We provide an algorithm for linear models in Section \ref{subsec:outlier}.
Linear models are popular because they are not as restrictive as they initially seem -- features can be transformed (e.g., by taking logs, square roots, etc.) so that they have an approximately linear relationship with the target.
The ideas from Section \ref{subsec:outlier} can be extended to more complex models, such as generalized linear models, and learned with variations of the expectation-maximization framework (EM) \cite{DEMP1977}.

Assuming records are independent, the expected number of outliers $K$ is the sum of the outlier probabilities of each record: $K=\lfloor\sum_{i=1}^n t_i\rfloor$. This means we can take the records with the top $K$ outlier probabilities and flag them as outliers. Since the system can accept many correlation templates as input, it will be learning many models, and a record is labeled as an outlier if any of these models flag it as an outlier. 

%)We aim at identifying outliers and give outlier probability as outlier score. %From our observation, there are many problematic records in the uncleaned dataset.
% For example, a taxi trip with trip time recorded as 4,294,966 seconds is considered anomalous because it might be unlikely for one taxi cab to drive for over a month. However,  we might not have any clue about whether it is normal for a trip with trip time as 2 hours. Therefore,  we look at other attributes (e.g., trip distance or fare) and consult with domain experts about what variables have correlations. Then we train a statistical model to learn how these variables are correlated. We define \emph{Outlier} as a data point which deviates from other observations in terms of given correlation.

\subsection{Outlier Data Modeling}\label{subsec:outlier}

For the purpose of simplicity and clearness, we use the following renaming in this section. For a correlation $(j, S)$, we set $y_i$ to be $\vec{z}_i[j]$ and $\vec{x}_i$ to be the vector of the attribute values in $S$ (i.e. $\vec{x}_i = [\vec{z}_i[s] \text{ for } s \in S]$).

Linear models have a weight vector $\vec{w}$, a noise random variable $\epsilon_i$, and the functional form
%{\small
\begin{equation}
 y_i = \vec{w} \cdot \vec{x}_i + \epsilon_i 
\label{eq:model}
\end{equation}
%}
%\vspace{-10pt}
The noise distribution $\epsilon_i$ for record $i$ is modeled as follows. We assume that there is a probability $p$ that a data point is an outlier. Hence, the error $\epsilon_i$ is modeled as a mixture distribution --  with probability $1-p$ it is a zero mean Gaussian with unknown variance $\sigma^2$, and with probability $p$ it is a Cauchy random variable.  Note that the Gaussian distribution has probability density 
%\vspace{-3pt}

%$f_G(\epsilon_i; {\sigma}^2) = \frac{1}{\sqrt{2\pi {\sigma}^2} } \; \text{exp} ( -\frac{\epsilon_i^2}{2\sigma^2} ).$

%{\small
\[ 
f_G(\epsilon_i; {\sigma}^2) = \frac{1}{\sqrt{2\pi {\sigma}^2} } \; \text{exp} ( -\frac{\epsilon_i^2}{2\sigma^2} )
\]
%}

%If all of our results use the Cauchy distribution, there is no reason to talk about the t distribution first, just change everything to Cauchy
%%%%On the other hand, the outliers are modeled by nonnegative mean and large variance \footnote{a sample $z$ from t-distribution with parameters $\mu_1$, $\eta$, and $\lambda$ is modeled as: 1) sample $\tau$ from the Gamma(a, b) distribution;  2) sample $z$ from Gaussian($\mu_1$, 1/$\tau$) distribution, where $\eta = 2a $ and $\lambda = a/b$.} with probability density function:
%the outliers are modeled by nonnegative mean and large variance t-distribution with parameters $\mu_1$, $\eta$, and $\lambda$ as follows: 
%how we get t distribution:
%\begin{itemize}
%\item  sample $\tau$ from the Gamma(a, b) distribution
%\[ \tau \sim \text{Gamma}(\tau; a, b) = \frac{b^{a} \tau^{a-1} e^{-\tau b}}{\Gamma(a)}  \]
%\item sample $z$ from Gaussian($\mu_1$, 1/$\tau$) distribution.
%\end{itemize}
%where $\eta = 2a $ and $\lambda = a/b$. Thus, the probability density function of t-distribution is

%\[ f_T(x; a, b, \mu_1, \tau) = \frac{b^a}{\Gamma(a)} {\tau_i}^{a-1} e^{-b\tau_i} \frac{\sqrt{\tau_i}}{\sqrt{2 \pi}} \text{exp} ( -\frac{ \tau_i( x - \mu_1)^2}{2}) \]

%\vspace{-5pt}
%Cauchy - mean and variance not exist. undefined. mu_1 is not the mean of cauchy!
The Cauchy distribution with scale parameter $b$ is a heavy-tailed distribution with undefined mean and variance, hence it is ideal for modeling outliers. It is equivalent to the Student's t distribution with 1 degree of freedom \cite{lange1989robust}. 

A sample $\epsilon_i$ from this distribution can be obtained by first sampling a value $\tau_i$ from the Gamma(0.5, b) distribution then sampling $\epsilon_i$ from the Gaussian($0$, 1/$\tau_i$) distribution \cite{balakrishnan2009continuous}. The probability of this joint sampling is
%{\small
\[ f_{C}(\epsilon_i, \tau_i;  b) = \frac{b^{0.5}}{\Gamma(0.5)} {\tau_i}^{0.5-1} e^{-b\tau_i} \frac{\sqrt{\tau_i}}{\sqrt{2 \pi}} \text{exp} ( -\frac{ \tau_i \epsilon_i^2}{2}) \]
%}
Given $\vec{x}_i$ and $y_i$, we introduce a latent indicator $\chi_i$ to denote where the error of $\vec{x}_i$ comes:
%{\small
\begin{align*}
\chi_i &= \begin{cases}
      1 & \text{ if the error of $\vec{x}_i$ is generated from the Cauchy}\\
      0 & \text{ if the error of $\vec{x}_i$ is generated from the Gaussian}
     \end{cases}
\end{align*}
%}
The expected value of $\chi_i$ is denoted by $t_i$ and is automatically computed by the EM algorithm.
With the model parameters $\vec{w}$ and unknown noise parameters $\sigma^2$ (variance of non-outliers), $p$ (outlier probability), $b$ (scale parameter of outlier distribution), the likelihood function is: 
%we use maximum-likelihood method to find proper parameter values for $\vec{w}$, $\sigma^2$, $p$ and $b$ is updated according to the interquartile estimate of scale for the Cauchy distribution. The likelihood function is: 
%{\small
\begin{equation}
\begin{aligned}
&L(\vec{w}, {\sigma}^2, p, b, \vec{\chi}, \vec{\tau})\\%&L(\vec{w}, {\sigma}^2, p, b, \vec{t}, \vec{\tau})\\
&=\prod_{i=1}^{n} \left[  (1-p) \frac{1}{\sqrt{2\pi {\sigma}^2} } \; \text{exp} ( -\frac{( y_i - \vec{w} \cdot \vec{x_i})^2}{2\sigma^2} )\right]^{1-\chi_i} \times \\ 
&\left[ p \frac{b^{0.5}}{\Gamma(0.5)} {\tau_i}^{0.5-1} e^{-b\tau_i} \frac{\sqrt{\tau_i}}{\sqrt{2 \pi}} \text{exp} ( -\frac{ \tau_i( y_i - \vec{w} \cdot \vec{x_i} )^2}{2} )  \right]^{\chi_i}  \\ 
\label{eq:likelihood}
\end{aligned}
\end{equation}
%}

\vspace{-5pt}
We iteratively update the estimates of $\sigma^2$, $p$, $b$, $\tau_i$ and  $t_i$  (the expected value of $\chi_i$) using the EM framework as described below. Note that the scale parameter $b$ of the Cauchy distribution cannot be estimated using maximum likelihood, so we update it using the interquartile range (the standard technique for Cauchy \cite{arnold2000skew}) as explained below.

%where $\mu_1 = 0$.

%\begin{comment}
%\begin{eqnarray*}
%\lefteqn{ L(\vec{w}, {\sigma}^2, p,  \mu_1, b, \vec{t}, \vec{\tau})}\\
%&\hspace{-0.3cm}=&\hspace{-0.3cm}  \prod_{i=1}^{n} \left[  (1-p) \frac{1}{\sqrt{2\pi {\sigma}^2} } \; \text{exp} ( -\frac{( y_i - \vec{w} \cdot \vec{x_i})^2}{2\sigma^2} )\right]^{1-t_i} \times \\ 
%&\hspace{-0.3cm} &\hspace{-0.3cm} \left[ p \frac{b^a}{\Gamma(\alpha)} {\tau_i}^{\alpha-1} e^{-b\tau_i} \frac{\sqrt{\tau_i}}{\sqrt{2 \pi}} \text{exp} ( -\frac{ \tau_i( y_i - \vec{w} \cdot \vec{x_i} - \mu_1)^2}{2} )  \right]^{t_i}  \\ 
%\end{eqnarray*}
%\end{comment}

 %Note that $a$ is set to 0.5, which specializes the distribution from Student t distribution to Cauchy distribution. The objective likelihood function is defined as 
%\begin{comment}
%\begin{equation} %this one does not plug in a
%\begin{aligned}
%&L(\vec{w}, {\sigma}^2, p,  \mu_1, b, \vec{t}, \vec{\tau})\\
%&=\prod_{i=1}^{n} \left[  (1-p) \frac{1}{\sqrt{2\pi {\sigma}^2} } \; \text{exp} ( -\frac{( y_i - \vec{w} \cdot \vec{x_i})^2}{2\sigma^2} )\right]^{1-t_i} \times \\ 
%&\left[ p \frac{b^a}{\Gamma(a)} {\tau_i}^{a-1} e^{-b\tau_i} \frac{\sqrt{\tau_i}}{\sqrt{2 \pi}} \text{exp} ( -\frac{ \tau_i( y_i - \vec{w} \cdot \vec{x_i} - \mu_1)^2}{2} )  \right]^{t_i}  \\ 
%\label{eq:likelihood}
%\end{aligned}
%\end{equation}
%\end{comment}

\subsection{Model Parameters Learning}

We employ EM algorithm \cite{DEMP1977} to solve the above likelihood function $L$. %The pseudo-code is shown in Algorithm \ref{alg:EM}.   
We iteratively update parameters so we add a superscript $^{(k)}$ to parameters to denote their values at the $k^{\text{th}}$ iteration. The E and M steps are described next.

\subsubsection{E step}

\begin{itemize}%[leftmargin=*,topsep=2pt,itemsep=1pt] %[leftmargin=*] %reduce indent
\setlength\itemsep{0em}
\item $\tau_i$ update: %Grouping the terms of $e$ in \cref{eq:likelihood}, 
 
 In \cref{eq:likelihood}, $\tau_i$ only appears in $e^{-\tau_i(b+0.5(y_i - \vec{w} \cdot \vec{x_i} )^2)}$ (after cancellation), which shows that $\tau_i$ (conditioned on the rest of the variables) follows exponential distribution. The conditional expected value of $\tau_i$ is

% $E(\tau_i) = \frac{1}{ b+0.5(y_i - \vec{w} \cdot \vec{x_i} )^2 }. $

%{\small
\begin{equation*}
 %E(\tau_i) =
 \frac{1}{ b+0.5(y_i - \vec{w} \cdot \vec{x_i} )^2 } 
\end{equation*}
%}

%After plugging in E($\tau_i$), 
By replacing $\tau_i$ with this expectation, the likelihood function $L$ in \cref{eq:likelihood}  is reduced to 
%{\small
\begin{equation}
\begin{aligned}
&L(\vec{w}, {\sigma}^2, p, b, \vec{\chi} ) \\
&= \prod_{i=1}^{n} \left[  (1-p) \frac{1}{\sqrt{2\pi {\sigma}^2} } \; \text{exp} ( -\frac{( y_i - \vec{w} \cdot \vec{x_i})^2}{2\sigma^2} )\right]^{1-\chi_i}  \times \\ 
& \left[ p \frac{\sqrt{b}}{\sqrt{\pi}} e^{-1} \frac{1}{\sqrt{2 \pi}}   \right]^{\chi_i} 
\label{eq:red_likelihood}
\end{aligned}
\end{equation}
%}
%\[ L(\vec{w}, {\sigma}^2, p,  \mu_1, b, \vec{t} ) =   \prod_{i=1}^{n} \left[  (1-p) \frac{1}{\sqrt{2\pi {\sigma}^2} } \; \text{exp} ( -\frac{( y_i - \vec{w} \cdot \vec{x_i})^2}{2\sigma^2} )\right]^{1-t_i}  \left[ p \frac{\sqrt{b}}{\sqrt{\pi}} e^{-1} \frac{1}{\sqrt{2 \pi}} )  \right]^{t_i}  \]

\smallskip

\item $t_i$ update (here $\text{sigmoid}(z) = \frac{1}{1+e^{-z}}$): 
%For each  $\vec{x}_i$, the inference procedure provides a number $t_i$ as an estimate of the probability that $\vec{x}_i$ is an outlier. 
%$t_i$ is updated according to the following equation . 
{\small
\begin{equation}
\begin{aligned}
%\hspace{-0.8cm}&t_i^{(k+1)} =\\
%\hspace{-0.8cm}& \text{sigmoid} ( \text{log}(\frac{p^{(k)}}{1-p^{(k)}})+ 0.5 \text{log}(\frac{b^{(k)} {\sigma^{2}}^{(k)}}{\pi e^2}) +
%\frac{ (y_i -\vec{w}^{(k)} \cdot \vec{x_i} )^2 }{2 {\sigma^{2}}^{(k)}} ) 
\hspace{-0.3cm}&t_i^{(k+1)} =\\
\hspace{-0.3cm}& \text{sigmoid} ( \text{log}(\frac{p^{(k)}}{1-p^{(k)}})+ 0.5 \text{log}(\frac{b^{(k)} {\sigma^{2}}^{(k)}}{\pi e^2}) +
\frac{ (y_i -\vec{w}^{(k)} \cdot \vec{x_i} )^2 }{2 {\sigma^{2}}^{(k)}} ) 
\end{aligned}
\label{eq:ti}
\end{equation}
}
\smallskip

%\hspace{-0.9cm} 
 %and $b = \frac{1}{Median(\vec{\xi})}$ \footnote{We have 3 versions of $b$ update: 1) $b$ is fixed to a constant $\pi e^2$, 2) choose $b$ so that the term $0.5 \text{log}(\frac{b \sigma^2}{\pi e^2})$ in $t_i$ update in equation \ref{eq:ti} is a non-positive constant -8,  3) $b$ = 1/(median errors of outliers). We apply method 3) because of lower prediction error.}. Note that $\vec{\xi}$ is the vector of absolute error $| y_i - \vec{w} \cdot \vec{x_i} | $ of each outlier data $x_i$.  Note that $b$ update is based on the interquartile estimate of scale for the Cauchy distribution.

\item $b$ update:  $b^{(k+1)} = \frac{1}{Median(\vec{\xi})}$ where $\vec{\xi}$ is the vector of absolute error $| y_i - \vec{w}^{(k)} \cdot \vec{x_i} | $ for the top $K$ records with highest $t_i^{(k)}$ values (note $K=\lfloor\sum_{j=1}^n t_j^{(k)}\rfloor$).

%\item $b$ update\footnote{We have 3 versions of $b$ update, but we use the $b$ = 1/ median errors of outliers for the model.}: Note that we tune the parameter $b$  in three approaches. First, $b$ is fixed to a constant $\pi e^2$. Second, we choose $b$ so that the term $0.5 \text{log}(\frac{b \sigma^2}{\pi e^2})$ in $t_i$ update in equation \ref{eq:ti} is a non-positive constant -8. This gives the data point $x_i$ with error ($y_i - \vec{w} \cdot \vec{x_i}$) larger than $4 \sigma$ more chance to be the outlier. Third, $b$ = 1/ median errors of outliers. This approach is based on the interquartile estimate of scale for the Cauchy distribution. Note that, for each iteration, the top ($p \times$ total number of records) data point with largest $t_i$ are labeled as outliers. The third solution to the $b$ update is applied for $t_i$ update. This decision is made because the third solution gives lower prediction error as compared to others.
\end{itemize}

%%%%%The M step is using E step's updated values (k+1) !!!

\subsubsection{M step}
 
 For each iteration before convergence, we update the estimated outlier probability $p$, the variance of non-outliers $\sigma^2$, and the coefficients $\vec{w}$. The updated parameters are listed below.
\begin{itemize}%[leftmargin=*,topsep=2pt,itemsep=1pt] 
\item $p$ update: $\quad$ 
 \begin{equation*}
%{\small
%$
p^{(k+1)}= \frac{1}{n} \sum_{i=1}^n {t_i}^{(k+1)}
%$
\end{equation*}
%}

\smallskip

\item $\sigma^2$ update:
%{\small
\begin{equation*}
%$ {\sigma^{2}}^{(k+1)} = \frac{\sum_{i=1}^n (1-{t_i}^{(k+1)})(y_i - \vec{w}^{(k)} \cdot \vec{x_i})^2}{n- \sum_{i=1}^{n} {t_i}^{(k+1)}}$
{\sigma^{2}}^{(k+1)} = \frac{\sum_{i=1}^n (1-{t_i}^{(k+1)})(y_i - \vec{w}^{(k)} \cdot \vec{x_i})^2}{n- \sum_{i=1}^{n} {t_i}^{(k+1)}}
\end{equation*}
%}

\smallskip

\item $\vec{w}$ update: %Let the coefficients be a $j$ dimension vector $\vec{w} = (\vec{w}[0], \vec{w}[1], \cdots, \vec{w}[j])$. 
$\vec{w}^{(k+1)}$ is the solution to the weighted least square problem where we give each ($y_i$, $\vec{x_i}$) a weight (1-$t_i^{(k+1)}$). Specifically, this weight (1-$t_i^{(k+1)}$) tells how much the model should rely on this datum. Thus, if one example has its $t_i^{(k+1)}$ as 1 (with probability 1 as an outlier), then it does not contribute to the $\vec{w}^{(k+1)}$ update coefficients. 

%The weighted least square estimation is
%\begin{align*}
%\begin{matrix}
%    (1-t_1)\\
%    (1-t_2)\\ 
%    \cdot \\
%     \cdot \\
%     \cdot \\
%     (1-t_n)\\
%\end{matrix}
%  \begin{bmatrix}
%     y_1\\
%     y_2\\
%     \cdot \\
%     \cdot \\
%     \cdot \\
%     y_n\\
%     \end{bmatrix}   
%=  
%  \begin{bmatrix}
%     \vec{x_1}\\
%     \vec{x_2}\\
%    \cdot \\
%     \cdot \\
%     \cdot \\
%      \vec{x_n}\\
%     \end{bmatrix} 
%  \begin{bmatrix}
%     \vec{w}[1]\\
%     \vec{w}[2]\\
%     \cdot \\
%     \cdot \\
%     \cdot \\
%      \vec{w}[j]\\
%     \end{bmatrix}    
%\end{align*}

%Therefore, the updated coefficients $\vec{w}^{(k+1)}$ is the solution to the equation

%Let $\vec{1}$ be the vector whose components are all 1. Therefore, 
The update for $\vec{w}^{(k+1)}$ is a weighted least squares update:
%{\small
\begin{equation*}
\vec{w}^{(k+1)}  \leftarrow  \left[{X^{\star}}^TX^{\star}\right]^{-1} {X^{\star}}^T  \vec{y^\star}
\end{equation*}
%}
where $X^{\star} = VX$, $\vec{y^\star} = V\vec{y}$ and $V$ is a $n$ by $n$ diagonal matrix with $V_{ii} = \sqrt{1-t_i^{(k+1)}}$.

% \begin{equation*}
 %X^{T} X \vec{w}^{(k+1)} = X^{T} (1- \vec{t }^{  (k) })^T \vec{y}
% \vec{w}^{(k+1)} \leftarrow  \left[X^TX\right]^{-1} X^T (1-\vec{t}^{ (k) })^T \vec{y}
 
%\end{equation*}

\end{itemize}

The algorithm will terminate when parameters $\vec{w}, \sigma, p, b$ converge. Since each iteration involves finding the median absolute error in $b$ update, the time complexity is $O(n \log n \cdot T)$ where $T$ is the number of iterations.

\subsection{Outlier Labeling}   
    
%   Outlier detection process contains a sequence of filters where each filter represents a given correlation. %Each filter is a trained linear regression model of equation \ref{eq:model}. 
  %how is each filter chosen?    
As mentioned before, every filter model assigns to every record $i$ a score $t_i$ indicating an estimated probability that it is an outlier and an estimated fraction of outliers $p$. The filter then labels a record an outlier if it has one of the top $K$ values of $t_i$ where $K=\lfloor\sum_{i=1}^n t_i\rfloor \approx p\times n$. %We do this for every filter model and a point is finally marked an outlier if any filter marks it as an outlier.

\smallskip

 %In this framework (flow shown in \ref{fig:flow}), for a given dataset, we filter it through $n$ filters sequentially.
 Each filter model identifies different types of outliers. After the data pass through $i$ filters, each record receives $i$ labels $\ell_1, \ell_2, \cdots \ell_i$ from $i$ filters where $\ell_i = 1$ indicates it is an outlier flagged by filter $i$ and $\ell_i = 0$ otherwise. At the end, we add this record to the outlier set if  $\ell_1 \vee \ell_2 \vee \cdots \vee \ell_i = 1$.

%% file: 6experiment_detector.tex
\section{Experiments}
\label{sec:exp}

The outlier detector was implemented using MapReduce. The rest of experiments used a machine with 2.00GHz Intel(R) Xeon(R) CPU and 48 GB RAM.

\subsection{Datasets}
  We apply our filtering model to four real-world unlabeled datasets.  We assume that records in ElNino and Houses datasets are not corrupted which is also an assumption in ~\cite{valko2011conditional,liang2016robust}, so we inject synthetic outliers. Other datasets such as Bodyfat \cite{bodyfat} and Algae \cite{algae}  used in \cite{valko2011conditional} exhibit correlations between attributes. However, we do not consider them in our experiments because the data size is too small.
  
\subsubsection{NYC Taxi}
%14087 true medallion
A large-scale 22GB public New York City taxi dataset \cite{nyctaxi} is collected from more than 14,000 taxis, which contains $173,179,771$ taxi trips from 01/01/2013 to 12/31/2013. Each record is a trip with attributes: medallion number (anonymized), hack license (anonymized), vendor, rate  code, store and forward flag, pickup location, pickup datetime, drop off location, dropoff datetime, passenger count, payment type, trip time, trip distance, fare amount, tips, tax, tolls, surcharge, and total amount. We use the subset of 143,540,889 trips which are within the Manhattan borough (the boundary is queried from \url{wikimapia.org}). We examine the outlying behavior in the trip time, distance, and fare.
%also rate code 1

\subsubsection{Intel Lab Sensor}
This is a public Intel sensor dataset  \cite{intelsensor} containing a log of about 2.3 million readings from 54 sensors deployed in the Intel Berkeley Research lab between 02/28/2004 to 04/05/2004. Each record is a sensor reading with date, time, sequence number,  sensor id, temperature ($\degree$C), humidity, light, voltage, and the coordinates of sensors' location. We consider two behavioral attributes as humidity and temperature. 

\subsubsection{ElNino}
 This dataset is from UCI repository \cite{elnino} with 93,935 records  after removing records with missing values. These readings are collected from buoys positioned around equatorial Pacific. The sea surface temperature is used as behavior variable while the rest of the oceanographic and meteorological variables are contextual variables.  

\subsubsection{Houses}
This dataset is from UCI repository  \cite{houses} with 20,640 observations on the housing in California. The house price is used as behavioral attributes and other variables such as median income, housing median age, total rooms, etc. are contextual attributes.

\vspace{-40pt}
\subsection{Initial Parameters Setting and Sensitivity}%{Filters' Initial Parameters Setting}
\label{sec:noise:param}

We describe how we decide the initial value for $p$, $b$, $\sigma^2$, and $\vec{w}$ used in the outlier detector.

\begin{figure}[t]
\centering
\subfigure[parameter $p$ sensitivity. \label{fig:sensitivity_p} ]{\includegraphics[height=2.5cm,width=7.3cm]{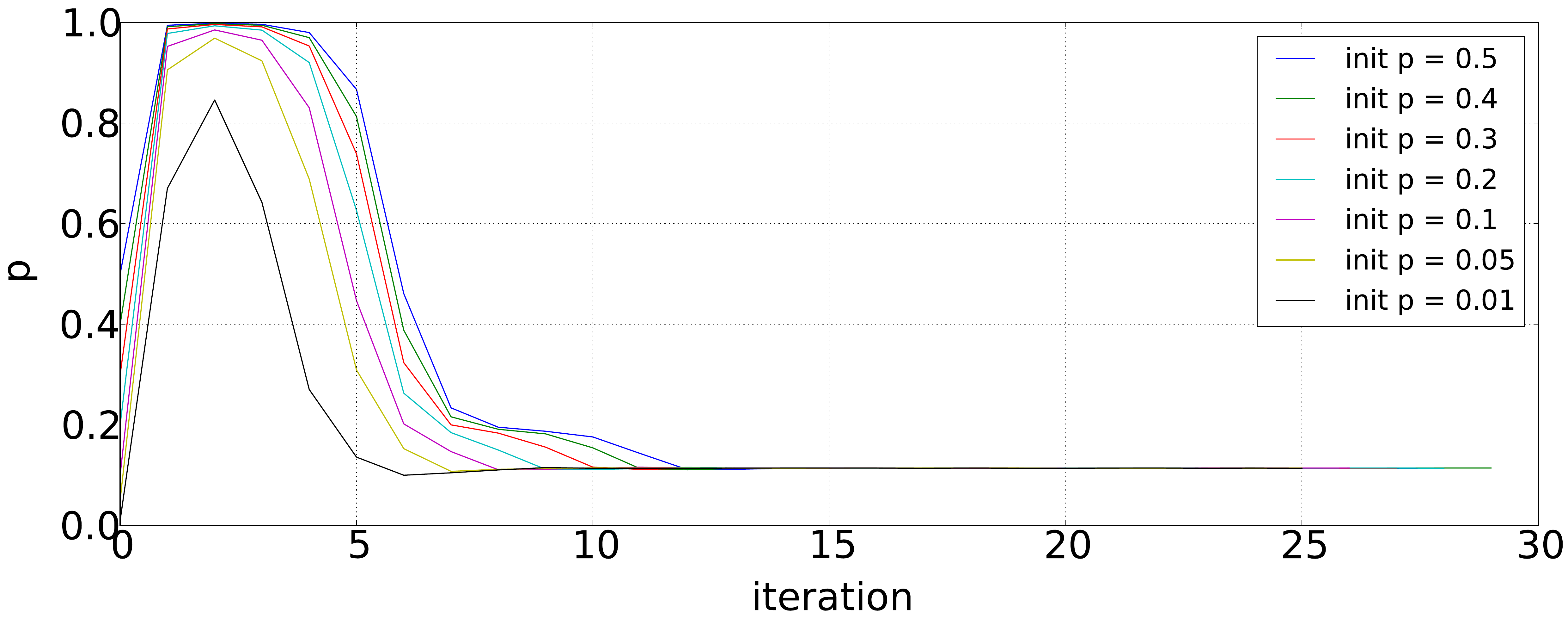}}%{figs/init_p_wide_font.pdf} }

\subfigure[parameter $\sigma^2$ sensitivity.\label{fig:sensitivity_sigma} ]{\includegraphics[height=2.5cm,width=7.3cm]{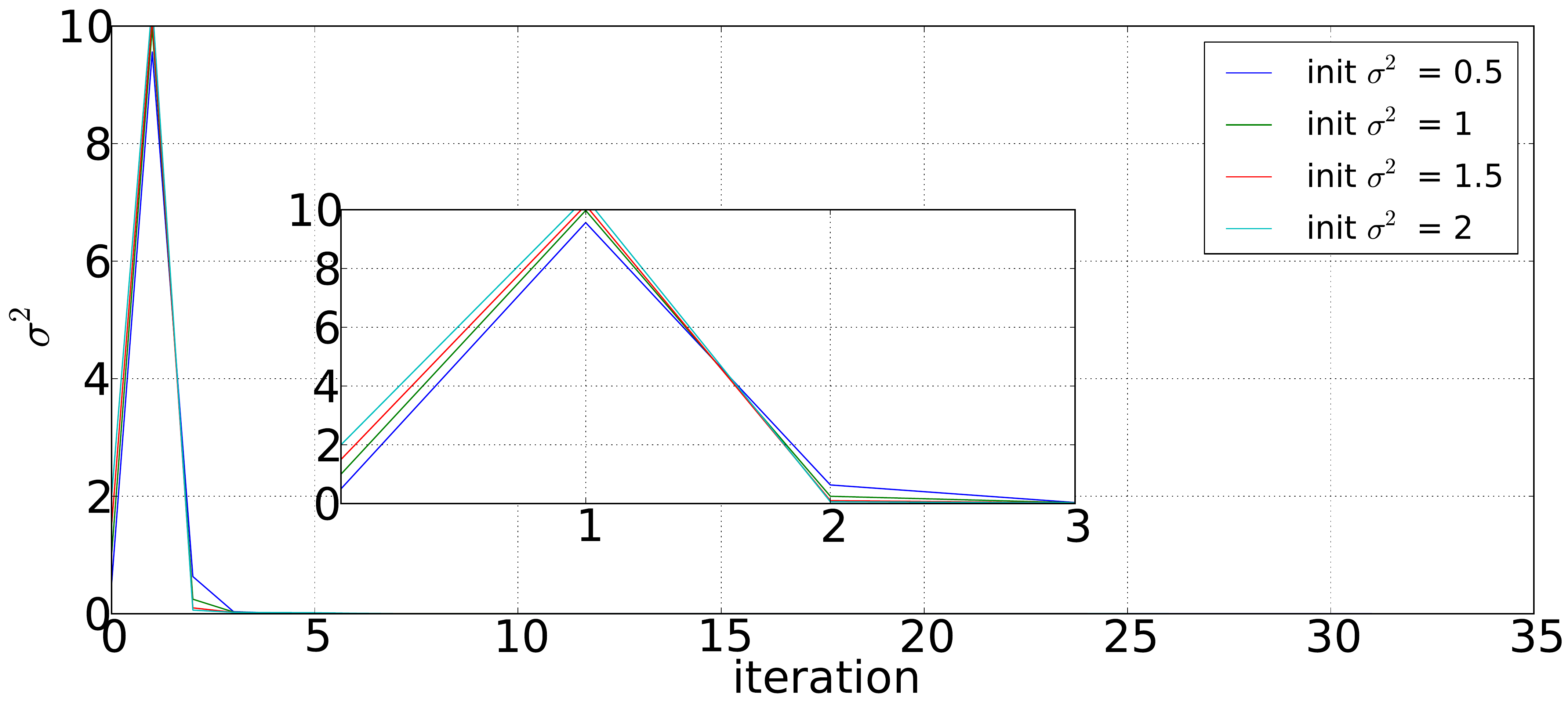}}%{figs/init_sigma_wide_font.png}}
\vspace{-10pt}
\caption{Parameter sensitivity of one filter used for sensor dataset.}
\label{fig:sensitivity}
\end{figure}

\begin{itemize}[leftmargin=*,topsep=2pt,itemsep=1pt] 
\item $p = 0.05$. %Recall that $p$ is an estimate of percentage of outliers in the entire dataset and 
We start with 5\% as initial value. We observed that the final converged value is not very sensitive  to initial settings.  Figure \ref{fig:sensitivity_p} gives an example of how $p$ changes in the iterations on NYC taxi dataset, with different starting points, to converge to approximately the same value 0.114. 

\smallskip

\item ${\sigma^2} = 1$. As ${\sigma^2}$ represents the variance of non-outliers, it is preferred to initially be a small number. In NYC taxi data, Figure \ref{fig:sensitivity_sigma} shows the convergence path of $\sigma^2$ with different starting values in $ [0.5, 2]$. 
%converges to the same value 0.001.

\item $\vec{w} = (1, 0, \cdots, 0)$. Let the feature variable be a $j$ dimension vector $\vec{x} = (\vec{x}[1], \vec{x}[2], \cdots, \vec{x}[j])$ and the target variable $y$. Suppose there is a linear relationship between variables $\vec{x}[1]$ and $y$.  The initial coefficient for $\vec{x}[1]$ is set to be 1, i.e.,  $\vec{w}[1] = 1$. Others are initialized as 0.

\smallskip
\item $b = \pi e^2$. The $b$ value only affects $t_i$. We choose this setting so that the term $0.5 \text{log}(\frac{b \sigma^2}{\pi e^2})$ in the $t_i$ update in Equation \ref{eq:ti} equals 0, and thus each data point's $t_i$ value in the beginning of the algorithm is dominated by the error ($y_i - \vec{w} \cdot \vec{x}_i$). 
\end{itemize}

\subsection{Outlier Detection Baselines}
\label{sec:baselines}
We evaluate \SysName against the state-of-the-art algorithms including traditional outlier detection, contextual outlier detection, regression models and methods specifically designed for outliers in taxi data.

\begin{itemize}[leftmargin=*,topsep=2pt,itemsep=1pt] 
\item  \textbf{density-based method}. A widely referenced density-based algorithm LOF \cite{breunig2000lof} outlier mining. %for global (i.e., full-space) outlier mining. 
We implement this method and adopt the commonly used settings for neighbor parameter $k = 10$.

%\smallskip
\item \textbf{distance-based method}. A recent distance-based outlier detection algorithm with sampling \cite{sugiyama2013rapid}. We use the provided code and the default sample size $s = 20$.

%\smallskip
\item \textbf{OLS}. The linear regression with ordinary least square estimation. The outlier score of record $i$ is its Cook's distance $\mathcal{D}_i$ = $\frac{e_{i}^2}{s^{2}p} \left[  \frac{h_{i}}{(1-h_{i})^2}   \right]$ where $e_i$ is the error of the $i$th record, $s^{2}$ is the mean squared error of the ordinary linear regression model, $p$ is the dimension of feature vector $\vec{x_i}$ and the leverage of record $i$ is $h_i = \vec{x}_i  [X^T X]^{-1} {\vec{x}_i}^{~T}$. 

%\smallskip
\item \textbf{GBT}. The gradient boosting tree regression model \cite{friedman2001greedy}. We select parameters from a validation set. The outlier score is defined as the absolute difference between the predicted value and the true value.

%\smallskip
\item \textbf{CAD \cite{song2007conditional}}. Conditional Anomaly Detection. A Gaussian mixture model $U$ is used to model contextual attributes $x$ where $U_i$ denotes the $i$-th component. Another Gaussian mixture model $V$ is used to model behavioral attributes $y$ with $V_j$. Then, a mapping function $p(V_j | U_i)$ is used to compute the probability of $V_j$ being selected under the condition that its contextual variables are generated from $U_j$. We set the number of Gaussian components as 30. The outlier score is defined as an inverse of the probability computed from this approach.

%\smallskip
\item \textbf{ROCOD \cite{liang2016robust}}. Robust Contextual Outlier Detection. An ensemble of local expected behavior and global expected behavior is used to detect outliers. For the local behavior, a neighbor-based locality sensitive hashing is used to locate contextual neighbors and an average of neighbors' behavior attribute is considered as expected local estimation. A linear ridge regression or non-linear tree regression is chosen to model the global expected behavior. We chose the non-linear model as its global estimation because it is the best performance in Houses and Elnino datasets reported in their work. The outlier score is computed as the absolute value of a weighted average of global and local estimates minus the true value. 

%\smallskip
\item \textbf{SOD \cite{zhang2012smarter}}. Smarter Outlier Detection. A method specifically designed for taxi dataset. SOD works by snapping the pickup and dropoff locations to the nearest street segments. The trips which fail to be mapped to the street are considered type I outliers. Next, it computes the shortest path distance and compares that to actual trip distance to detect outliers (called type II outliers). It is worth noting that our outlier filtering model can also employ road network for detecting outliers by simply using the shortest path distance as an input feature. However, we do not do this so that we can give SOD an advantage, while seeing how other features in the data can be used to detect anomalies.

%\item \textbf{statistical-based method}. A baseline we designed for taxi dataset. Since we observe some detour trips in the taxi data. We fit the ratio of travel distance and L2 distance between end points into Gaussian distribution. The outlier score of point $x$ is $1-p(x)$ where $p(x)$ is the gaussian density function.

\end{itemize}

%\subsubsection{Intel Sensor}
\subsection{Experiments on Intel Sensor Data}%\subsection{Outlier Evaluation on Intel Sensor}
%DESCCRIBE THE OUTLIERS CLAIMED IN Scorpion
%number of sensor, number of outliers claimed in Scorpion (manually), our automatically approaches , 
We describe the filters and confirm detected outliers with Scorpion \cite{wu2013scorpion}, which also uses the sensor data for evaluation.
%\textbf{Sensor filters}. 
\subsubsection{Sensor filters}
In this dataset, the temperature is correlated with voltage and humidity. We use 2 filters below. The first filter marks  5.2\% of total records as outliers ($p = 0.052$) while the second filter marks 11.4\% ($p = 0.114$). Among these marked records, 44\%  are captured by both filters. Note that sensor's age refers to the days or weeks since these sensors were deployed.
%%%%Note that sensor's age obviously improves the number of outliers detected. Take filter 2 as example, $p$ increases from 0.0047 to 0.114 when we add sensor's age as features. 
\begin{enumerate}[leftmargin=*,topsep=2pt,itemsep=1pt] %\topsep: space between first item and preceding paragraph. \itemsep: space between successive items.
\item log(humidity) = $w_{1}$ $\times$ log(temperature) + $\vec{w_a} \cdot \vec{a_d}$  + $\beta_1$, where $\vec{a_d}$ is sensor's age measured in days.

\item log(temperature) = $w_{2}$ $\times$ log(voltage) + $\vec{w_a'} \cdot \vec{a_w}$  + $\beta_1'$, where $\vec{a_w}$ is sensor's age measured in weeks.
\end{enumerate}

%3) Section 5.4.2 spends a long time explaining the outliers. Isn't that the job of the rules? Can the rules explain the outliers? How different are they from your manual analysis? I think this discussion should be merged with the rule discussion. The scorpion section should flow like: description of outliers discussed in scorpion paper, statistic about outliers detected by SysName ) -- how many outliers from each method, what is the overlap. Then explain how you plan to validate these outliers (machine learning), present the results. Then introduce the rules discussion ("Now let us examine the interpretation of the outliers provided by SysName"), and then include the merged discussion of rules and what you had in 5.4.2.

\subsubsection{Method for Comparison}%\subsubsection{Sensor Outlier Evaluation}
 \label{sec:sensor_baselines}

Because the dataset does not contain ground truth, we validate our detected Intel sensor outliers with findings of Wu and Madden in the Scorpion system \cite{wu2013scorpion} where they, using domain knowledge, manually identify one type of outliers.
 
 %flow1: description of outliers discussed in scorpion paper, statistic about outliers detected by SysName ) -- how many outliers from each method, what is the overlap
 The problematic sensors claimed in Scorpion are temperature readings $\in$ (90\degree C, 122\degree C) generated from sensor 15 and sensor 18 and they account for 5.6\% of records in the whole dataset.  Approximately 11\% of records are flagged by our system  \SysName, including  manually identified outliers in the Scorpion paper. While those manual annotations provide some ground truth (i.e. have high precision), they may not have flagged all outliers (i.e. recall is unknown).%We discuss the rest of outliers (5.4\%) that is not mentioned in Scorpion as follows.

 %91860 out of total 2219803 (~0.0413) sensor records are outliers detected by ordinary least square
 We also compare with linear regression model with ordinary least squares estimatation (OLS). We apply the Cook's distance ($\mathcal{D}$) to estimate the influence, or the combination of leverage and residual values, of each record. Points with large Cook's distance are considered to have further examination. We flag outliers as points with $\mathcal{D} > 4/n$ where n is the number of observations  \cite{bollen1985regression}. The result shows that 4.13\% of records are flagged as outliers by OLS. We do not choose other outlier detection methods listed in \Cref{sec:baselines} because none of them provides a threshold in outlier score for users to flag outliers.
 
 %flow2: Then explain how you plan to validate these outliers (machine learning), present the results
 %validate by Use the different outlier methods to modify training data, build model on modified training data, then test on testing set.
 %1. "outlier methods can only be used on training data, cannot look at testing data"
%2. you can either train on all training data, or training data minus outliers, or use weighted training data where weights are 1-prob_data_point_is_outlier (some methods like weighted least squares regression can use point weights)

\subsubsection{Evaluation Metric}
 We validate flagged outliers by machine learning tasks. In cases where ground truth is missing, it is customary to divide data into training/testing sets. We run our outlier detector on the training set and use the flagged training records to modify the training data (i.e., remove or downweight records suspected of being outliers). Then we build various machine learning models on the modified training data. The goal is to compare these accuracy of the models on a common testing set. The main intuition is that uncaught outliers will degrade the training of the models and thus hurt testing accuracy; better outlier detection algorithms are therefore more likely to result in good training datasets that yield models to perform better on testing data.

To follow this intuition, we use 5-fold cross validation and design prediction tasks with linear and non-linear regression models. The evaluation metrics are mean absolute error (MAE) and mean relative error (MRE). Since there are also anomalous records in the testing data, we also use the median absolute error (MedAE) and median relative error (MedRE).
In \cref{tab:prediction_tasks}, we employ linear regression (LR), support vector regression with quadratic error function (SVR), and decision tree regression (DTR) and we put all attributes as features for the prediction task1 -- predicting temperature where temperature is the variable involved in 2 outlier detecting filters. We train the models on four different training sets -- all training set, all training set minus Scorpion outliers, all training set minus \SysName outliers, and all weighted training set. %Note that we use the geometric mean for the outlier score from two filters as the weight of a data point. 
Note that we use the scikit-learn \cite{scikit-learn} implementation for the model LR, SVR and DTR.

\subsubsection{Results}
Results are presented in \cref{tab:prediction_tasks}. In Task1 with models LR and SVR, removing our detected outliers from training set or down-weighting those outliers results in lower error. We also conduct the Paired Student's t-test and Wilcoxon signed rank test to show that it is statistically significant that the MAE of our modified training set (i.e., training set minus our detected outliers) is lower than the MAE of Scorpion's modified training set.

Note that with the DTR model,  using all the training data gets the lower \emph{mean} absolute and relative error. However, because the testing data does contain outliers, the mean can be skewed by them. The median errors (MedAE, MedRE) are more robust measures of performance and  show that taking out outliers in training set (or downweighting them) leads to more accurate prediction. %for non-outliers. %Comparing the mean error in last two columns, outliers in the testing set makes the results less reliable because the errors of the outlying trips greatly impact the average accuracy.

\begin{table*}%[b]%this table should appear at the bottom of the page.
[t]%it is placed here in the input to force its position at the top of the next page.
\vspace{-15pt}
\caption{Performance with/without Outlier}
\centering
%\begin{tabular}{|c||c|c||c|c||c|c|}
%\begin{tabular}{|c||c|c||c|c|}
\begin{tabular}{|c||c|c||c|c||c|c|c|c|||c|c|c|c|}
\hline
%  \multicolumn{5}{|c|}{Prediction Task 1 }   \\ \hline
 % & \multicolumn{2}{|c||}{ 1} & \multicolumn{2}{|c||}{ 2} & \multicolumn{2}{|c|}{ 3}  \\ \hline
%%Model & \multicolumn{2}{c||}{LR} & \multicolumn{2}{c|}{SVR}    \\ 
%%%Model & \multicolumn{2}{c||}{LR} & \multicolumn{2}{c||}{SVR (quadratic error)}  & \multicolumn{2}{c|}{DTR}    \\ 
Model & \multicolumn{2}{c||}{LR} & \multicolumn{2}{c||}{SVR}  & \multicolumn{4}{c|||}{DTR} & \multicolumn{4}{c|}{DTR (test set without outlier)}    \\ 
\hline
Train Set & MAE & MRE & MAE      & MRE & MAE      & MRE & MedAE & MedRE &  MAE & MRE  & MedAE & MedRE  \\ 
 \hline
 \hline
% \multicolumn{5}{|c|}{Task: temperature prediction based on 2 filters result }   \\ \hline
% \multicolumn{7}{|c|}{Task1: temperature prediction based on 2 filters result }   \\ \hline
\multicolumn{13}{|c|}{Task1: temperature prediction}   \\ \hline
all train & 11.94 &0.66 & 14.10 & 0.75 & \bf{3.53} & \bf{0.21} & 1.12 & 0.05 & 2.15 & 0.1 &1.13 & 0.05   \\ 
 \hline
- Scorpion & 11.94 &0.66 & 11.23 & 0.65 & 3.54 & 0.21 & 1.12 & 0.05 & 2.15 & 0.1 & 1.14 & 0.05 \\ 
 \hline
 - OLS & 9.8 &0.6 & 14.83 & 0.74 & 3.75 & 0.38 & 1.1 & 0.05 & 1.88 & 0.09 & 1.11 & 0.05 \\ 
 \hline
- $\SysName$ & \bf{9.56} & \bf{0.53} & 9.49 & 0.54 & 5.7 & 0.36 & \bf{0.9} & \bf{0.04} & \bf{1.18} & \bf{0.05} & \bf{0.87} & \bf{0.04}  \\ 
 \hline
weighted & \bf{9.56}  & \bf{0.53} & \bf{6.91} & \bf{0.46} & 5.66 & 0.37 & \bf{0.9} & \bf{0.04} & \bf{1.19} & \bf{0.05} & \bf{0.87} &\bf{0.04} \\
 \hline
 %%\hline
%% \multicolumn{13}{|c|}{Task2: voltage prediction based on filter1 result }   \\ \hline
%%  all train & 0.085 &0.034 & 0.15 & 0.061 & 0.068 & 0.028 & 0.059 & 0.024 & 0.069 & 0.028 & 0.062 & 0.025  \\ 
%% \hline
%%  - Scorpion & 0.085 &0.034 & 0.09 & 0.038  & 0.068 & 0.027 & 0.059 & 0.023 & 0.068 & 0.028 & 0.059 & 0.023  \\ 
%% \hline
%% - $\SysName$ & 0.092 & 0.038 & 0.09 & 0.038 & 0.07& 0.029 & 0.058 & 0.024 & 0.069 & 0.028 & 0.061 & 0.024  \\ 
%% \hline
%% weighted  &0.092 & 0.038&  0.62 & 0.254 & 0.07 & 0.03 & 0.063 & 0.026 & 0.069 & 0.028 & 0.06 & 0.024\\
%%\hline
%%\hline
%% \multicolumn{13}{|c|}{Task3: temperature prediction based on filter1 result }   \\ \hline
%%  all train & 11.97 & 0.65 & 12.2 & 0.7 & \bf{3.46} & \bf{0.21} & 1.09 & 0.05 & 2.33 & 0.1 & 1.05 & 0.05 \\ 
%% \hline
%%  - Scorpion & 11.97 & 0.65 & 20.32 & 0.88 & \bf{3.46} & \bf{0.21} & 1.09 & 0.05 & 2.33 & 0.1 & 1.05 & 0.05  \\ 
%% \hline
%% - $\SysName$ & \bf{9.42} & \bf{0.55} & 9.47 & 0.54 & 5.29 & 0.36 & 1.05 & 0.04 & 1.36 & 0.06 & 0.97 & 0.04 \\ 
 %%\hline
%% weighted & \bf{9.42} & \bf{0.55} & \bf{6.91} & \bf{0.45} & 5.7 & 0.37 & \bf{0.91} & \bf{0.04} & \bf{1.32} & \bf{0.06} & \bf{0.83} & \bf{0.04} \\
%%\hline
\end{tabular}
\label{tab:prediction_tasks}
\end{table*}

\subsubsection{Case Study} 
\begin{figure}
 %\hspace{-3pt}
	\centering
	\includegraphics[width=7.1cm]{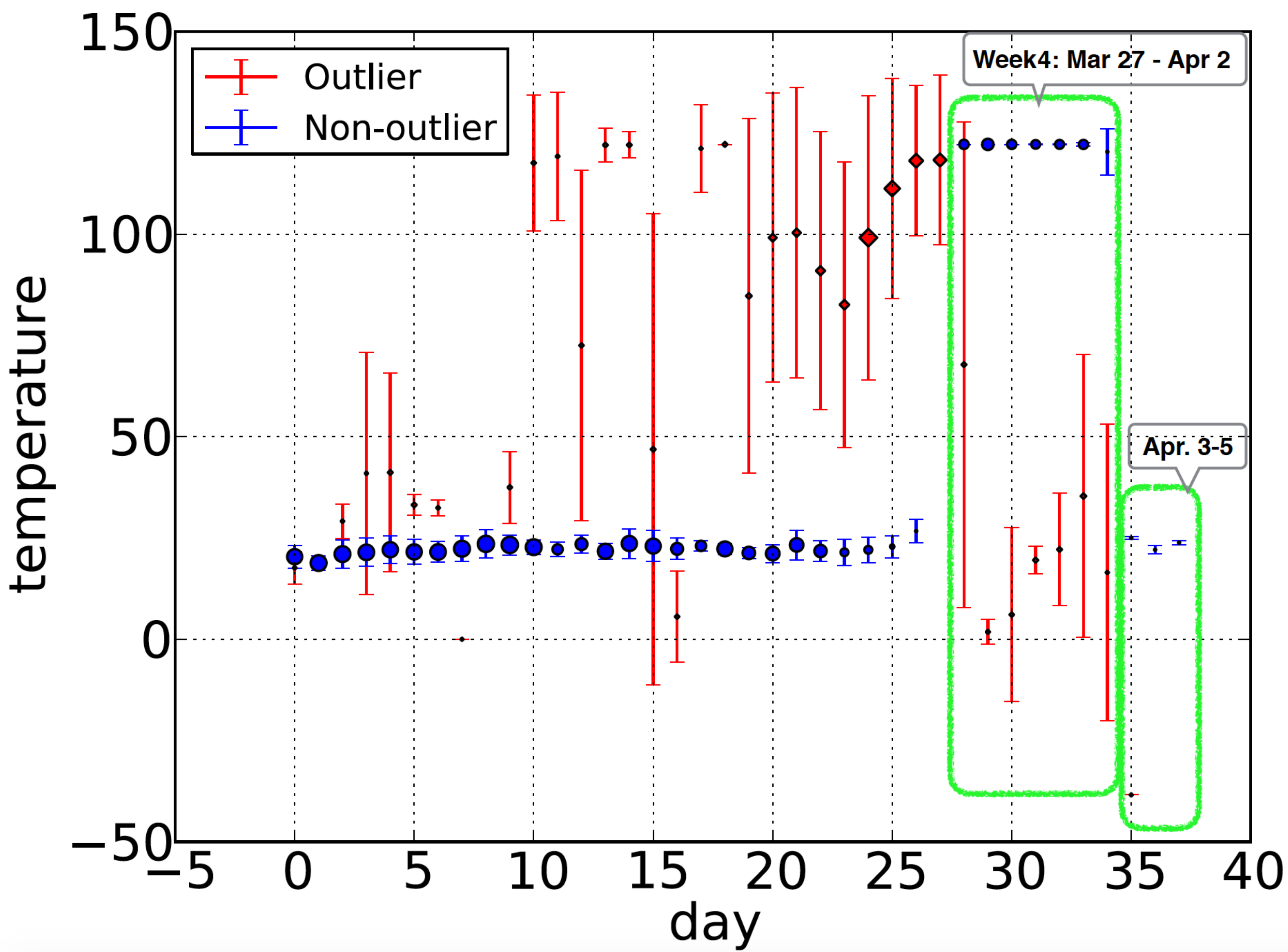}
	\vspace{-5pt}
	\caption{The average temperature readings sequence  from 2/28/2004 to 4/5/2004}
\label{fig:temp}
\end{figure}

 57.2\% of flagged outliers are associated with anomalous temperature reading in Week 3. \SysName observes a general sensor's malfunction pattern  as it is unlikely to be real temperature in the lab -- fifty five out of total fifty eight sensors exhibit that the temperature reading is increasing until it reaches around 122\degree C and it keeps generating 122\degree C or above in Week 3 (as shown in \Cref{fig:temp}). However, on Week 4, almost all sensors generate temperature $\in$ [122.15, 175.68) -- hence that is the norm in the data for that week. This is a very common pattern in the data that 92\% of records produced on week 4 generates temperature $\in$ [122.15, 175.68). Hence they are classified as normal by \SysName. Note that Scorpion refers to sensors generating high temperature as problematic sensors.

\begin{figure}
%  \hspace{-3pt}
	\centering
	\includegraphics[width=7.1cm]{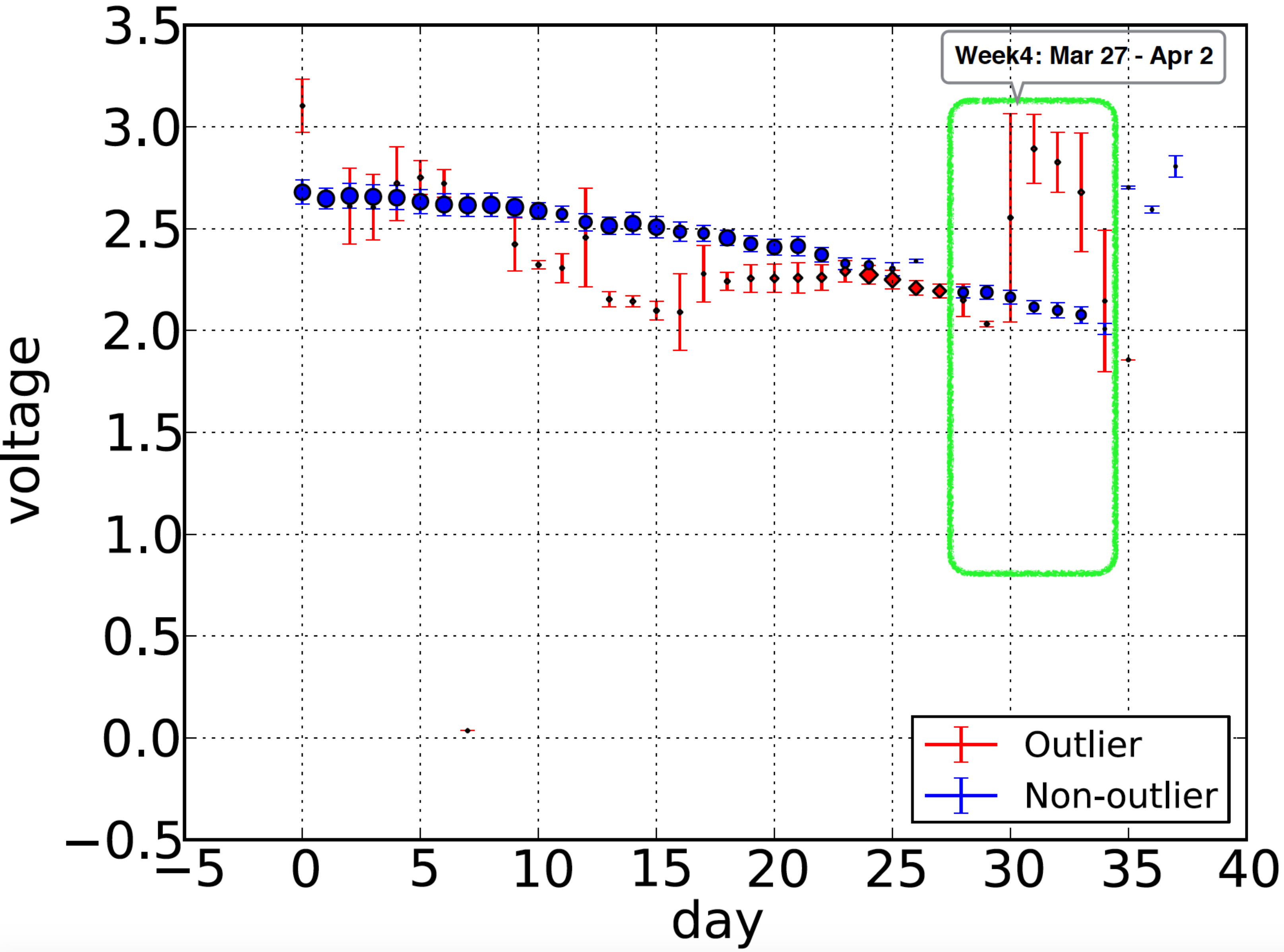}
	\vspace{-5pt}
	\caption{The average voltage readings sequence from 2/28/2004 to 4/5/2004}
\label{fig:sensor_v}
\end{figure}

  In figure \ref{fig:sensor_v}, we see that there is a decreasing trend in voltage for this batch of sensors and this helps to justify the fact that records with voltage $\geq$ 2.8 in Week 4 are identified as outliers by \SysName.
 
\subsection{Experiments on NYC Taxi}%\subsection{Outlier Evaluation in NYC Taxi}   

First we describe five filters we used to detect outliers. We validate the results with human-annotated trips and compare with a method called Smarter Outlier Detection (SOD) \cite{zhang2012smarter}, which was specifically designed for this dataset. In Section \ref{sec:existing_outlier_detectors} we also compare against the state of the art  outlier detection algorithms.
%SOD works by snapping the pickup and dropoff locations to the nearest street segments. The trips which fail to be mapped to the street are considered type I outliers. Next, SOD computes the shortest path distance and compares it to actual trip distance to detect outliers (called type II outliers). It is worth noting that our outlier filtering model can also employ road network for detecting outliers by simply using the shortest path distance as an input feature. However, we do not do this so that we can give SOD an advantage, while seeing how other features in the data can be used to detect anomalies.

\subsubsection{Taxi filters}
 %\textbf{Taxi filters}. 
We used the following filtering models (where $w_i$ is the coefficient and $\beta_i$ is offset).   
\begin{enumerate}[leftmargin=*,topsep=2pt,itemsep=0.5pt] 
\item Trip time = $w_{1}$$\times$(dropoff time-pickup time)+$\beta_1$ \label{item:1}

\item Fare = $w_2$$\times$(total amount-tips-tax-toll-surcharge)+$\beta_2$ \label{item:2}

\item log(Trip time) = $w_3$$\times$log(Fare)+$\beta_3$   \label{item:3} %w = 1.3240975283070213,3.4095643959072675

\item log(Trip distance) = $w_4$$\times$log(L2)+$\beta_4$   \label{item:4}  %w = 0.9531233637292201,0.26577390650513477 from 20150213 

\item log(Trip time) = $w_5$ $\times$ log(L2) + $\vec{w_t} \cdot \vec{ts} + \beta_5$, where  $\vec{ts}$ is the vector of 24-dimension temporal features described below.  \label{item:5}
\end{enumerate} 
 
 %The parameters of filters are shown in table \ref{tab:filter-details}. 
\smallskip
Filter \ref{item:1} and \ref{item:2} encode what should be functional dependencies. However, they may differ due to software bugs, data entry errors, or device miscalibration. For example, trip time might be recorded by the taxi meter while pickup and drop-off times might be recorded by a gps unit with a separate clock. In Filters  \ref{item:3}, \ref{item:4}, \ref{item:5}, trip time/distance/fare/L2-displacement are all positively correlated and we expect their variance to grow proportionally with the length of a trip. For this reason, we use logs (so that multiplicative error becomes additive error). Also, trip time may depend on the time of day (e.g., rush hour), so we include those components in Filter \ref{item:5}. 
%We use the L2 distance between two end points to predict trip distance for filter \ref{item:4} and predict trip time for filter \ref{item:5}. 
%Since adding the location features for pickup and dropoff location pair such as partition the manhattan into grids does not improve the prediction, we only use the L2 distance in filter \ref{item:4}. Also, trip distance is not influenced by the time of day. In contrast, we observe the same trip time patterns during the weekdays but different patterns on weekends. 
We partition time of day into 2-hour time slots and separate out weekends from weekdays (this giving $12 \times 2=24$ temporal features). We note that filters \ref{item:3} and \ref{item:5} did not have overlap in the records they flagged as outliers, thus showing that correlated sensor readings can fail in different ways.

\subsubsection{Taxi Outlier Detection Method Comparison}

    %We design a human labeling system for volunteers/experienced taxi riders to determine outlier trips and to provide reasons to support their judgement. Note that we give the rate of fare information from Taxi \& Limousine Commission on the labeling system so that people are able to get the rough idea to check the trip attributes. Each trip is labeled by three people and we take the majority votes as the ground truth. Then, we compare the NYC taxi outlier detection results with SOD.
    
    In total, \SysName flagged 7\% of records as outliers (the dataset is known to be noisy).  
    The code for SOD was not available,  so we reproduced it  with a different software package.\footnote{\url{http://project-osrm.org}} We discarded trips whose end points are not on roads. The dataset can be categorized into  four disjoint sets: $\boldsymbol{M\widetilde{S}}$ - records flagged as outliers by \SysName but not SOD,  $\boldsymbol{\widetilde{M}S}$ - records flagged as outliers by SOD but not \SysName, $\boldsymbol{MS}$ - records flagged as outliers by both methods, and $\boldsymbol{\widetilde{M}\widetilde{S}}$  - records not flagged by any method.
 
 %note:
 %$\boldsymbol{My\widetilde{S}}$ is S1
 %$\boldsymbol{\widetilde{My}S}$ is S2
 %$\boldsymbol{MyS}$  is S3
 %$\boldsymbol{\widetilde{My}\widetilde{S}}$ is S4
    
\subsubsection{Evaluation Metric}
\label{sec:taxi_outlier} 
 We designed a human labeling system for experienced taxi riders to determine outlier trips and to provide reasons to support their judgements. We provided the labelers with the taxi fare rate information from the NYC Taxi \& Limousine Commission. Each trip is labeled by three people and we take the majority votes as the ground truth. 
 
    To provide a quantitative evaluation, each time the labeling webpage randomly selects 10 trips from each of the 4 sets for a person to label. 
    
 \subsubsection{Results} 
   In all, 6517 trips were labeled and the results are shown in \cref{tab:4sets}. In set $\boldsymbol{M\widetilde{S}}$, 92\% of trips were labeled by humans as outliers and these are consistent with our approach. %However, 8\% of trips are labeled as non-outliers and taken as non-outlier by SOD. According to the reasons collected from human labeling system, some people think of the round trip as non-outlier, which contradicts with our filtering model. Thus, our accuracy of $\boldsymbol{My\widetilde{S}}$ could improve if the round trips are seen as outlier. 
   
   In set  $\boldsymbol{\widetilde{M}S}$, humans only labeled 16\% of the records as outliers. Thus, when \SysName and SOD disagreed, humans tended to agree with the classification provided by $\SysName$.
   %labeled trips are real non-outliers. They are also identified as non-outliers by our method whereas detected as outliers by SOD. The rest 16\% of trips are labeled outliers because most of the trips whose travel distance is smaller than the straight line distance. One possible reason that we fail to catch these trips is the approximation error from haversine distance between longitudes and latitudes. Clearly, our method outperforms SOD in $\boldsymbol{My\widetilde{S}}$ and  $\boldsymbol{\widetilde{My}S}$.
   For $\boldsymbol{MS}$ and $\boldsymbol{\widetilde{M}\widetilde{S}}$, both our method and SOD get the accuracy of 98\% and 94\% respectively. 

\vspace{-5pt}
\begin{table}[htb] %h here , t:top , b, bottom
\center
\caption{4 Sets of Labeled Trips}
\vspace{-4pt}
\begin{tabular}{ | C{1.2cm} | C{1.5cm} | C{1.4cm} | c |  }
%\toprule  %fraction of trips in $Set_i$ means for all the trips, there are 6.6% of trips which is classified in Set_i
\hline
\\[-1em] 
    $Set_i$    & \% trips in $Set_i$  &  \# labeled trips  &  $\frac{\text{\# labeled outliers}}{\text{\# labeled trips in $Set_i$}}$ \\ %[2ex]
\hline  
\\[-1em] 
  $\boldsymbol{M\widetilde{S}}$ & 6.6\% & 1739 & 92\% \\%\textbf{S1} - $O_{my}G_{SOD}$ & 6.6\% & 1739 & 92\% \\
 \hline
 \\[-1em] 
$\boldsymbol{\widetilde{M}S}$ & 0.093\% & 1698 & 16\% \\%\textbf{S2} - $G_{my}O_{SOD}$ & 0.093\% & 1698 & 16\% \\
 \hline
  \\[-1em] 
$\boldsymbol{MS}$  & 0.416\% & 1570 & 98\% \\% \textbf{S3} - $O_{my}O_{SOD}$ & 0.416\% & 1570 & 98\% \\
 \hline
  \\[-1em] 
$\boldsymbol{\widetilde{M}\widetilde{S}}$  & 90.27\% & 1510 & 6\% \\%\textbf{S4} - $G_{my}G_{SOD}$ & 90.27\% & 1510 & 6\% \\
 \hline
%\bottomrule
\end{tabular}
\label{tab:4sets}
\end{table}

%The evaluation criteria of the overall outlier detection performance are detection rate ($DR$, i.e., the fraction of outliers that are successfully detected as outliers), false positive rate ($FPR$, i.e., the fraction of normal records that are predicted to be outlier), precision (i.e., fraction of detected outlier that are real outlier), true negative rate ($TNR$, i.e., fraction of non-outlier that are detected as non-outlier), and accuracy. According to table \ref{tab:contingency}, 
%\begin{equation*}
% DR = \frac{TP}{TP+FN}, ~~ FPR = \frac{FP}{FP+TN}
%\end{equation*}
%\begin{equation*}
% Precision = \frac{TP}{TP+FP}, ~~ TNR = \frac{TN}{TN+FP}
%\end{equation*}
%\begin{equation*}
% Accuracy = \frac{TP+TN}{TP+FP+FN+TN}
%\end{equation*}

%According to \cref{tab:contingency}, 
We  use the following evaluation criteria for overall outlier detection performance: detection rate ($DR = \frac{TP}{TP+FN}$, i.e., fraction of outliers that are successfully detected as outliers), false positive rate ($FPR = \frac{FP}{FP+TN}$, i.e.,  fraction of normal records that are predicted to be outlier), precision ($Precision = \frac{TP}{TP+FP}$, i.e., fraction of detected outlier that are real outlier), true negative rate ($TNR = \frac{TN}{TN+FP}$, i.e., fraction of non-outlier that are detected as non-outlier).%, and $ Accuracy =$ $\frac{TP+TN}{TP+FP+FN+TN}$.

Note that we use the labeled sampled trips to estimate the ground truth statistics for entire dataset. The estimation approach is as follow: suppose, in set $i$, the total number of trips is $u$; the number of sampled trips labeled is $v$; out of labeled trips $v$ the outliers account for $\phi$ \%. Hence $u \times \phi$ is the estimated outliers for set $i$. The evaluation on both the labeled trips (denoted as on labeled) and estimated results for all trips (denoted as on all) are presented in \cref{tab:detection_res}. From the results it is clear that \SysName achieves much better detection rates for slightly larger false positive rates.
\nop{
\begin{table}[htb] %h here , t:top , b, bottom
\center
\caption{Contingency table}
\begin{tabular}{ | C{2.8cm} | C{1.8cm} | C{1.8cm} |   }
%\toprule
\hline 
         & Labeled Outlier  & Labeled   Non-outliers    \\
\hline
Predict Outlier  &  TP & FP    \\ 
\hline
Predict  Non-outlier & FN  & TN  \\     
\hline
%\bottomrule
\end{tabular}
\label{tab:contingency}
\end{table}
}

\vspace{-14pt}
\begin{table}[htb] 
\center
\caption{Outlier Detection Performance}
\vspace{-5pt}
\begin{tabular}{ | c | c | c | c | c |  }
%\toprule 
\hline
         & \multicolumn{2}{c|}{Our method \SysName}  & \multicolumn{2}{c|}{Competitor SOD}     \\
 \cline{2-5}
   	&  on all & on labeled & on all & on labeled \\
\hline
  DR	&  \textbf{0.55} & \textbf{0.89} & 0.035 & 0.52 \\
\hline
  FPR	&  0.006 & \textbf{0.057} & \textbf{0.001} & 0.484 \\	
\hline
  Precision &  \textbf{0.924} & \textbf{0.94} & 0.83 & 0.55 \\
 \hline
  TNR	&  \textbf{0.99} & \textbf{0.942} & 0.99 & 0.515 \\
 \hline
%  Accuracy	&  \textbf{0.94} & \textbf{0.92} & 0.88 & 0.51 \\			
%\hline
%\bottomrule
\end{tabular}
\label{tab:detection_res}
\end{table}

\vspace{-10pt}
\subsubsection{Outlier Detection Methods Comparison}
\label{sec:existing_outlier_detectors}
We evaluate \SysName against traditional outlier detection and contextual outlier detection approaches described in \Cref{sec:baselines}. We also adopt the following statistical-based method as baseline.

%\smallskip
 \textbf{statistical-based method}. Since we observe some detour trips in the taxi data. We fit the ratio of travel distance and L2 distance between end points into Gaussian distribution. The outlier score of point $x$ is $1-p(x)$ where $p(x)$ is the gaussian density function.

%\vspace{-5pt}
\textbf{Evaluation metrics}. We randomly sampled 22,463 trips as input data to these outlier detectors which give the outlier rank to every trip. The labeling process uses the same type of methodology mentioned in \cref{sec:taxi_outlier}. Given the outlier rank of the trips, we first select every 5 trips to be labeled (i.e., $5^{th}$, $10^{th}$ $\dots$). Thus we get a rough idea of the approximate number of outliers in this sampled dataset. We label the top 1400 trips for each method and use the following metrics for evaluation.

\vspace{-5pt}
{\small
\begin{equation*}
 \text{Precision $@ \kappa$} = \frac{ \text{\# trips whose rank $\leq \kappa$ and label = Outlier} }{\kappa }
\end{equation*}
}

\begin{figure}
	\centering
	\includegraphics[width=7.2cm]{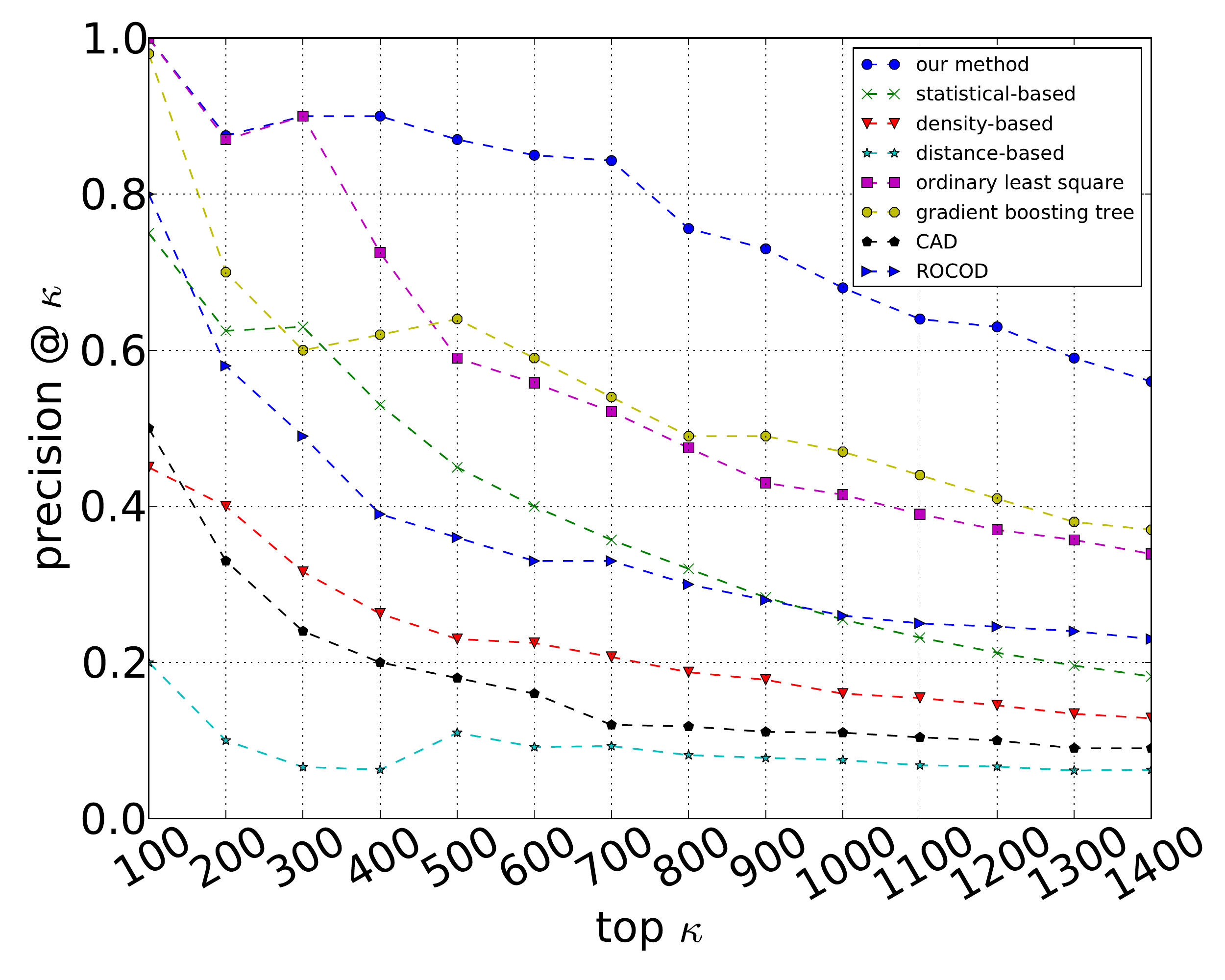}%{figs/precision_at_n_all_methods_frontsize.pdf}
	\vspace{-15pt}
	\caption{Precision at $\kappa$ }
\label{fig:precision_at_n}
\end{figure}

\vspace{-10pt}
\cref{fig:precision_at_n} shows that our method outperforms others. The top outlier trips detected by density-based and distance-based methods are long trips such as trips from upper to lower manhattan. Even though they have less neighbors than shorter trips, their trip information is considered as reasonable by the labeled outcome. In contrast, our top outlier trips are mainly from device error and thus it could be obviously identified by people. For the linear and non-linear regression model as well as the existing contextual outlier detection (CAD \& ROCOD),  they can identify extreme outliers in their top 100 outliers. But their precision drops with more false positives which is due to the biased prediction.  We find that trips with rank greater than 1200 are mostly labeled as non-outliers. Hence the precision $@ \kappa$ drops around $\kappa = 1000$.

\subsubsection{Case Study}
We describe the anomalous trips and interesting findings. First, we point out the payment systems and trip time tracking systems provided by Creative Mobile Technologies (CMT) are programmed differently from those provided by Verifone (VTS). We expect that the travel time should be consistent with the duration of dropoff time subtracted by pickup time. Our identified outliers show that this is not always the case and the discrepancies are almost always associated with vendor CMT. Similarly, the sum of cost fields (i.e.,Tip, Tax, Surcharge, Toll, Fare) should equal to total amount (Total). Those inconsistent cost related fields are all produced by the VTS payment system.

Second, we identify a type of outliers ranked lower as compared to those extreme corrupted records. These records contain trip fare $<$ \$3. The metered fare regulated by Taxi \& Limousine Commission (TLC) initially charges \$2.5 once a passenger gets in taxis and plus \$0.5 per 0.2 mile or \$0.5 per minute in slow traffic. We believe these records are outliers as they are even less than minimum taxi fare.

Last, 1\% of detected anomalous records are trips with almost the same GPS coordinates from pickup to dropoff location. To investigate this further, we found the park-cemetery manhattan neighborhood and bridges linking manhattan to nearby boroughs (where gps signal might be weak) are highly correlated with these trips.

\vspace{-3pt}
\subsection{Experiments on Synthetic Outlier Data}
\label{sec:synthdata}

   For the Elnino and Houses datasets, we inject synthetic outliers into the original clean data. %Unlike the perturbation scheme in \cite{liang2016robust,song2007conditional}, we explore different types of outliers where we give controls to where/how many outliers are injected or its degree of outlierness.
   One perturbation scheme used in \cite{liang2016robust,song2007conditional} is that they first randomly select a sample $\vec{z_i} = (\vec{x_i}, y_i)$ then, from $k$ data points of the entire dataset, select another sample $\vec{z_j} = (\vec{x_j}, y_j)$ where the difference between $y_i$ and $y_j$ is maximized. This new data point $(\vec{x_i}, y_j)$ is added as an outlier. We do not follow this scheme for several reasons. First, swapping the attribute values may not always obtain desired outliers. It is likely that most of the swaps could result in normal data. Second, as we observe many extreme outliers in the real-world datasets, swapping values between samples in a clean data is less likely to produce this extreme difference between $y_i$ and $y_j$. Here we present  another way to generate outliers and we explore different types of outliers where we give controls to where and how many outliers are injected or its degree of outlierness.
      
\subsubsection{Perturbation Scheme}  To inject $q \times N$ outliers into a dataset with $N$ data samples, we randomly select $q \times N$ records $\vec{z_i} = (\vec{x_i}, y_i)$ to be perturbed. Let $y_i$ be the target attribute for perturbation. Let $\vec{x_i}$ be the rest of attributes. For all selected records, a random number from (0, $\alpha$) is added up to $y_i$ as $y_i^{~\prime}$. Then we add new sample $\vec{z}^{~\prime} = (\vec{x_i}, y_i^{~\prime})$ into the original dataset and flag it as outlier. Note that original $N$ data samples are flagged as non-outlier. In the experiments, we standardized the target attribute to range (18, 30) which are the min and max value of the behavioral attribute in Elnino dataset. Set $\alpha$ as 50 by default. 

\subsubsection{Evaluation Metric}
Since all these outlier detection approaches considered in \Cref{sec:baselines} give rank to each record according to outlier score,  the Precision-Recall curve (PRC) is obtained by Precision $@ \kappa$ and Recall $@ \kappa$ for all possible $\kappa$ where the first $\kappa$ ranked records are determined to be outlier. The evaluation metric we use here is the Area Under the Curve (AUC) of the Precision-Recall curve instead of the Receiver Operating Characteristic (ROC) as it is less informative in imbalanced class problem \cite{davis2006relationship}.

\subsubsection{Results}
As up to 6\% of records in Sensor dataset are flagged as outliers due to sensor malfunction, we vary the perturbation ratio $q$ from 0.01 to 0.15 to see if our model is robust in the presence of a large fractions of anomalies. The performance in terms of AUC is shown in the following tables.% \Cref{tab:auc, tab:auc_contextual}, \Cref{tab:auc_contextual} and \Cref{tab:auc_size}. 

\Cref{tab:auc} presents the results when we perturb behavioral attributes to generate outliers. $\SysName$ consistently perform the best and its difference compared to other methods becomes significant when more outliers are involved ($q > 0.05$).
 Another type of synthetic outliers is produced by adding noise to contextual attributes. To see how it affects the performance, we select features with highest Pearson correlation to behavioral attribute for perturbation.  In \Cref{tab:auc_contextual}, we observe that a small fraction of outliers in contextual attribute could hurt the performance considerably for the other methods, especially for the tree-based approaches such as ROCOD and GBT on these two datasets. However, our method is robust and resistant to the fraction of outliers.
 
 We next investigate degree of outlierness of the injected anomalies. As $\alpha$ increases, larger magnitude of noise will have more chance to be added to the original value. Consequently, there are more extreme outliers and our performance is increased as expected in \Cref{tab:auc_size}. %In general, one can see that $\SysName$ constantly outperform others and outliers could break the effectiveness of existing algorithms.

\begin{table*}
\vspace{-10pt}
\centering
\caption{AUC of Precision Recall Curve w.r.t different fractions of synthetic outliers in behavioral attribute}
\vspace{-5pt}
\begin{tabular}{ | c | c | c | c | c | c | | c | c | c | c | c|}
%\toprule 
\hline
 &\multicolumn{5}{c||}{Elnino} & \multicolumn{5}{c|}{Houses}\\
\hline
   method         & q=0.01 & q=0.03  & q=0.05 & q=0.1 & q=0.15 & q=0.01 & q=0.03 & q=0.05 & q=0.1 & q=0.15 \\
\hline
\hline
\textbf{\SysName} & \textbf{0.96}& \textbf{0.97}  &  \textbf{0.98} & \textbf{0.98} & \textbf{0.98} & \textbf{0.93} & \textbf{0.92} &  \textbf{0.93} &  \textbf{0.95} &  \textbf{0.96}\\
\hline
\hline
ROCOD (non-linear) & 0.73 & 0.73 & 0.74 & 0.73 & 0.72 & 0.50 & 0.49 & 0.50 & 0.49  & 0.50  \\
\hline
CAD & 0.80 & 0.84 & 0.86 & 0.85 & 0.88 &  0.58 & 0.67 &  0.68 &  0.72 &  0.75\\
\hline
OLS & \textbf{0.96} & 0.95 & 0.95 & 0.92 & 0.90 & 0.92 & 0.91 & 0.92 & 0.91 & 0.91\\
\hline
GBT & \textbf{0.96} & 0.95 & 0.95 & 0.92 & 0.90 &  \textbf{0.93} & 0.91 & 0.92 & 0.91  & 0.91\\
\hline
distance-based & 0.81& 0.74 & 0.77 & 0.83 & 0.60 &  0.76 & 0.19 &  0.57 & 0.4 & 0.39 \\
\hline
density-based & 0.21  & 0.38 & 0.45 & 0.38 & 0.34 &  0.84 & 0.58 & 0.46 & 0.53 & 0.58\\
\hline
\end{tabular}
\label{tab:auc}
\end{table*}

%alpha = 50, x =  air.temp
\begin{table*}
\centering
\vspace{-8pt}
\caption{AUC of Precision Recall Curve w.r.t different fractions of synthetic outliers in contextual attribute}
\vspace{-8pt}
\begin{tabular}{ | c | c | c | c | c | c | | c | c | c | c | c|}
%\toprule 
\hline
 &\multicolumn{5}{c||}{Elnino} & \multicolumn{5}{c|}{Houses}\\
\hline
   method         & q=0.005 & q=0.01  & q=0.03 & q=0.05 & q=0.07 & q=0.005 & q=0.01 & q=0.03 & q=0.05 & q=0.07 \\
\hline
\hline
\textbf{\SysName} & \textbf{0.97}& \textbf{0.95}  &  \textbf{0.97} & \textbf{0.98} & \textbf{0.98} & \textbf{0.86} & \textbf{0.80} &  \textbf{0.88} &  \textbf{0.88} &  \textbf{0.91}\\
\hline
\hline
ROCOD (non-linear) & 0.01 & 0.01 & 0.02 & 0.02 & 0.03 & 0.03 & 0.01 & 0.02 & 0.04 & 0.05  \\
\hline
CAD & 0.80 & 0.83 & 0.86 & 0.88 & 0.87 &  0.51 & 0.54 &  0.56 &  0.61 &  0.63\\
\hline
OLS & 0.92 & 0.86 & 0.68 & 0.45 & 0.32 & 0.84 & 0.75 & 0.71 & 0.59 & 0.50\\
\hline
GBT & 0.11 & 0.15 & 0.28 & 0.37 & 0.40 & 0.04 & 0.04 & 0.08 & 0.11  & 0.15\\
\hline
distance-based & 0.88 & 0.74 & 0.81 & 0.50 & 0.83 &  0.54 & 0.73 &  0.22 & 0.20 & 0.42 \\
\hline
density-based & 0.08  & 0.07 & 0.08 & 0.09 & 0.10 &  0.01 & 0.01 & 0.03 & 0.04 & 0.06\\
\hline
\end{tabular}
\label{tab:auc_contextual}
\end{table*}

\vspace{-5pt}
%q = 0.01
\begin{table*}
\centering
\vspace{-8pt}
\caption{AUC of Precision Recall Curve w.r.t degree of outlierness $\alpha$ in contextual attribute}
\vspace{-7pt}
\begin{tabular}{ | c | c | c | c | c | c | | c | c | c | c | c|}
%\toprule 
\hline
 &\multicolumn{5}{c||}{Elnino} & \multicolumn{5}{c|}{Houses}\\
\hline
   method           &  $\alpha = 20$ & $\alpha = 30$ & $\alpha = 50$  &$\alpha = 100$ & $\alpha = 300$  & $\alpha = 30$ & $\alpha = 50$  &$\alpha = 100$ & $\alpha = 300$ & $\alpha = 500$ \\
\hline
\hline
\textbf{\SysName} &  \textbf{0.91} & \textbf{0.94}& \textbf{0.95}  &  \textbf{0.98} & \textbf{0.99}  & \textbf{0.75} & \textbf{0.8} &  \textbf{0.94} &  \textbf{0.97} &  \textbf{0.99}\\
\hline
\hline
ROCOD (non-linear) & 0.01 & 0.01 & 0.01 & 0.01 & 0.01  & 0.01 & 0.01 & 0.01 & 0.01 & 0.01  \\
\hline
CAD & 0.78 & 0.8 & 0.83 & 0.87 & 0.93 &   0.37 & 0.54 &  0.58 &  0.74 &  0.85\\
\hline
OLS & 0.88 & 0.89 & 0.86 & 0.85 & 0.73  & 0.72 & 0.75 & 0.87 & 0.86 & 0.83\\
\hline
GBT & 0.17 & 0.17 & 0.15 & 0.17 & 0.17  & 0.04 & 0.04 & 0.03 & 0.02  & 0.01\\
\hline
distance-based & 0.21 & 0.79 & 0.74 & 0.88 & 0.91  &  0.14 & 0.73 &  0.79 & 0.85 & 0.80 \\
\hline
density-based & 0.13 & 0.10  & 0.07 & 0.05 & 0.04  &  0.01 & 0.01 & 0.01 & 0.02 & 0.05\\
\hline
\end{tabular}
\label{tab:auc_size}
\end{table*}

%% file: 7conclusion_detector.tex
% !TEX root = cikm-main.tex

\section{Conclusions}
\label{sec:conclusion}
Motivated by a real-world problem, we develop a system \SysName which aims to detect outliers and explicitly considers outliers effect in modeling. It is a robust outlier detector as compared to the existing algorithms built on all the data records where their model parameters are skewed by outliers. Our method could potentially facilitate the public or research use of large-scale data collected from a network of sensors. 